\documentclass{article}
\usepackage{iclr2019_conference,times}

\usepackage[utf8]{inputenc}
\usepackage[T1]{fontenc}
\usepackage{hyperref}
\usepackage{url}
\usepackage{booktabs}
\usepackage{amsfonts}
\usepackage{nicefrac}
\usepackage{bbm}
\usepackage{microtype}
\usepackage{color}
\usepackage[]{graphicx}
\usepackage{amsmath}
\usepackage{xcolor}
\usepackage{paralist}
\usepackage{enumerate}
\usepackage[]{graphicx}
\usepackage{siunitx}
\usepackage{hyperref}
\usepackage{subfigure}
\usepackage{multicol}
\usepackage{wrapfig}
\usepackage{mathabx}

\usepackage{algorithmicx}
\usepackage{algpseudocode}
\usepackage{algorithm}

\usepackage{threeparttable}
\usepackage{tikz}
\usepackage{multirow}

\usepackage{array}
\newcolumntype{C}[1]{>{\centering\let\newline\\\arraybackslash\hspace{0pt}}m{#1}}

\usepackage{caption} 
\def\EE{\mathbb{E}}
\def\KL{\mathsf{KL}}

\def\EE{\mathbb{E}}
\def\MutI{\mathsf{MI}}

\definecolor{CommentOrange}{rgb}{0.8,0.4,0.0}

\newcommand{\Ent}{\mathsf{H}}

\newcommand{\qq}{\pi}
\newcommand{\pp}{\pi^0}
\newcommand{\traj}{\pi_{\tau}}
\newcommand{\xD}{x^\mathcal{D}}
\newcommand{\xG}{x^\mathcal{G}}

\title{Information asymmetry in KL-regularized RL}
\author{Alexandre Galashov, Siddhant M. Jayakumar, Leonard Hasenclever, Dhruva Tirumala,\\
\textbf{Jonathan Schwarz, Guillaume Desjardins, Wojciech M. Czarnecki, Yee Whye Teh}, \\
\textbf{Razvan Pascanu, Nicolas Heess}\\
DeepMind\\
London, UK\\
\texttt{\{agalashov,sidmj,leonardh,dhruvat,schwarzjn,gdesjardins,}\\
\texttt{lejlot,ywteh,razp,heess\}@google.com}
}

\date{}
\iclrfinalcopy 

\begin{document}

\maketitle

\begin{abstract}
    Many real world tasks exhibit rich structure that is repeated across different parts of the state space or in time. In this work we study the possibility of leveraging such repeated structure to speed up and regularize learning. We start from the KL regularized expected reward objective 
    which introduces an additional component, a default policy. Instead of relying on a fixed default policy, we learn it from data. But crucially, we restrict the amount of information the default policy receives, forcing it to learn reusable behaviours that 
    help the policy learn faster. 
    We 
    formalize this strategy and
    discuss connections to information bottleneck approaches and to the variational EM algorithm. We present empirical results in both discrete and continuous action domains and demonstrate that, for certain tasks, learning a default policy alongside the policy can significantly speed up and improve learning.
\end{abstract}

\section{Introduction}

For many interesting reinforcement learning tasks, good policies exhibit  
similar behaviors 
in different contexts, behaviors that need to be modified only  slightly or occasionally to account for the specific task at hand or to respond to  information becoming available. For example, a simulated humanoid in navigational tasks is usually required to walk -- independently of the specific goal it is aiming for. Similarly, an agent in a simulated maze tends to primarily move forward with occasional left/right turns at intersections.

This intuition has been explored across multiple fields, from cognitive science \citep[e.g.][]{wouter2018mental} to  neuroscience and machine learning. For instance, the idea of bounded rationality \cite[e.g.][]{simon1956rational} 
emphasizes the cost of information processing and the presence of internal computational constraints. This implies that the behavior of an agent 
minimizes the need to process information, and more generally trades off task reward with computational effort, resulting in structured repetitive patterns. Computationally, these ideas can be modeled using tools from information and probability theory \cite[e.g.][]{tishby2011information,ortega2011information,still2012information,rubin2012trading,ortega2013thermodynamics,tiomkin2017unified}, for instance, via constraints on the channel capacity between past states and future actions in a Markov decision process.

\begin{wrapfigure}{r}{0.45\textwidth}
\vspace*{-2em}
    \begin{center}
        \includegraphics[width=0.44\textwidth]{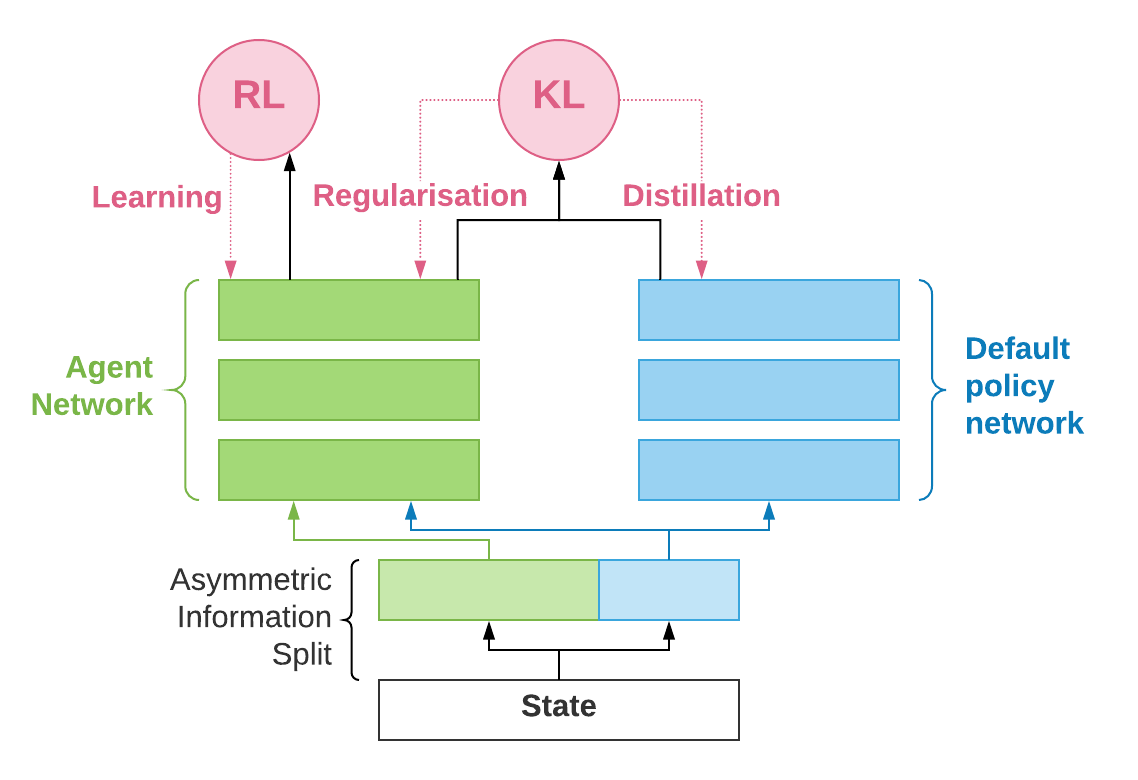}
        \caption{{Default policy-agent architecture}.}
        \label{fig:prior_posterior_architecture}
    \end{center}
\vspace*{-2em}
\end{wrapfigure}

In this paper we explore this idea, starting from the KL regularized expected reward objective \citep[e.g.][]{todorov2007linearly,toussaint2009robot,kappen2012optimal,rawlik12,levine2013variational,teh2017distral}, which encourages an agent to trade off expected reward against deviations from a prior or default distribution over trajectories. 
%
%
We explore how this can be used to inject subjective knowledge into the learning problem by using an informative default policy that is learned alongside the agent policy
This default policy encodes default behaviours that should be executed in multiple contexts in absence of additional task information and the objective
forces the learned policy to be structured in alignment with the default policy.

To render this approach effective, we introduce an information asymmetry between the default and agent policies, preventing the default policy from accessing certain information in the state.
This prevents the default policy from collapsing to the agent's policy. Instead, the default policy is forced to generalize across a subset of states, implementing a form of default behavior that is valid in the absence of the missing information, and thereby exerting pressure that encourages sharing of behavior across different parts of the state space.


Figure~\ref{fig:prior_posterior_architecture} illustrates the proposed setup, with asymmetry imposed by hiding parts of the 
state from the default policy.
We investigate the proposed approach empirically on a variety of challenging problems including both continuous action problems such as controlling simulated high-dimensional physical embodied agents, as well as discrete action visual mazes. 
%
%
We find that even when the agent and default policies are learned at the same time, significant speed-ups can be achieved on a range of tasks.
We consider several variations of the formulation, and discuss its connection to several ideas in the wider literature, including  information bottleneck, 
and variational formulations of the EM algorithm for learning generative models.



\section{KL and entropy regularized reinforcement learning}
\label{sec:KLregularized}

Throughout this paper we use $s_t$ and $a_t$ to denote the state and action at time step $t$, and $r(s,a)$ the instantaneous reward for the agent if it executes action $a$ in state $s$. We denote the history up to time $t$ by $x_t=(s_1,a_1,\ldots,s_t)$, and the whole trajectory by  $\tau=(s_1,a_1,s_2,\ldots)$. 
Our starting point is the KL regularized expected reward objective
\begin{align}
\textstyle
    \mathcal{L}(\qq, \pp) = \EE_{\traj} \left[ \sum_t  \gamma^t r(s_t,a_t) - \alpha
    \gamma^t \KL\left[ \qq(a_t|x_t) \| \pp(a_t | x_t)\right] \right], \label{eq:objective:standard}
\end{align}
%
where $\qq$ is the agent policy (parameterized by $\theta$ and to be learned), $\pp$ the default policy, and
$\EE_{\traj}[\cdot]$ is taken with respect to the  distribution $\traj$ over trajectories defined by the agent policy and system dynamics: $\traj(\tau) = p(s_1) \prod_t \qq(a_t | x_t) p(s_{t+1} | s_t, a_t)$. Note that our policies are history-dependent. 
$\KL[\qq(a_t|x_t) \| \pp(a_t | x_t)]$ is the Kullback-Leibler (KL) divergence between the agent policy $\qq$ and a default or prior policy $\pp$ given history $x_t$. The discount factor is $\gamma\in[0,1]$ and $\alpha$ is a hyperparameter scaling the relative contributions of both terms.

Intuitively, this objective expresses the desire to maximize the reward while also staying close to a reference behaviour defined by $\pp$.
As discussed later, besides being a convenient way to express a regularized RL problem, it also has deep connections to probabilistic inference.
One particular instantiation of eq.\ (\ref{eq:objective:standard}) is when $\pp$ is the uniform distribution (assuming a compact action space). 
In this case one recovers, up to a constant, the entropy regularized objective \citep[e.g.][]{ziebart2010modeling,fox2015taming,haarnoja2017reinforcement,schulman2017equivalence,hausman2018learning}:
\begin{align}
\textstyle
    \mathcal{L}_H(\qq) = \EE_{\traj}\left[ \sum_t  \gamma^t r(s_t,a_t) + \alpha\gamma^t 
\Ent[ \qq(a_t | x_t)] \right].
\label{eq:objective:entropy}
\end{align}
This objective has been motivated in various ways: it prevents the policy from collapsing to a deterministic solution thus improving exploration, it encourages learning of multiple solutions to a task which can facilitate transfer, and it provides robustness to perturbations and model mismatch. 
One approximation of the  entropy regularized objective is for the history dependent entropy to be used as an additional (auxiliary) loss to the RL loss; this approach is widely used in the literature \cite[e.g.][]{williams1991function,mnih2016asynchronous}.
While the motivations for considering the entropy regularized objective are intuitive and reasonable, the choice of regularizing towards an uniform policy is less obvious, particularly in cases with large or high dimensional action spaces. In this work we explore whether regularization towards more sophisticated default policies can be advantageous.

Both objectives \eqref{eq:objective:standard} and \eqref{eq:objective:entropy} can be generalized beyond the typical Markov assumption in MDPs. In particular, additional correlations among actions can be introduced, e.g.\ using latent variables \cite{hausman2018learning}.  This can be useful when, as discussed below, either $\pp$ or $\qq$ are not given full access to the state, rendering the setup partially observed. In the following we will not explore such extensions, though note that we do work with  policies $\qq(a_t|x_t)$ and $\pp(a_t|x_t)$ that depend on history $x_t$.

\section{Learning default policies}
\label{sec:LearnedPriors}

Many works that consider 
the KL regularized objective either employ a simple or fixed default policy 
or directly work with the entropy formulation \citep[e.g.][]{rubin2012trading,fox2015taming,haarnoja2017reinforcement,hausman2018learning}.
%
In contrast, here we will be studying the possibility of learning the default policy itself, and the form of the subjective knowledge that this introduces to the learning system.
%
%
Our guiding intuition, as described earlier, is the notion of a default behaviour that is executed in the absence of additional goal-directed information. Instances which we  explore in this paper include a locomotive body navigating to a goal location where the locomotion pattern depends largely on the body configuration and less so on the goal, and a 3D visual maze environment with discrete actions, where the typical action includes forward motion, regardless of the specific task at hand.




To express the notion of a default behavior, which we  also refer to as ``goal-agnostic'' (although the term should be understood very broadly), we consider the case where the default policy $\pp$ is a function 
(parameterized by $\phi$) of a subset of the interaction history up to time $t$, i.e.\ $\pp(a_t | x_t) = \pp(a_t | \xD_t)$, where $\xD_t$ is a subset of the full history $x_t$ and is the goal-agnostic information that we allow the default policy to depend on. We denote by $\xG_t$ the other (goal-directed) information in $x_t$ and assume that the full history is the disjoint union of both. The objective \eqref{eq:objective:standard} specializes to:
\begin{align}
\textstyle
    \mathcal{L}(\qq, \pp) = \EE_{\traj} \left[ \sum_t  \gamma^t r(s_t,a_t) - \alpha
    \gamma^t \KL\left[ \qq(a_t|x_t) \| \pp(a_t | \xD_t)\right] \right], \label{eq:objective:infoasym}
\end{align}
To give a few examples: If $\xD_t$ is empty then the default policy does not depend on the history at all (e.g.\ uniform policy). If $\xD_t=a_{1:t-1}$ then it depends only on past actions. 
%
In multitask learning $\xG_t$ can be the task identifier, while $\xD_t$ the state history. And finally, in continuous control $\xD_t$ can contain proprioceptive information about the body, while $\xG_t$ contains exteroceptive (goal-directed) information (e.g.\ vision). 

By hiding information from the default policy, the system forces the default policy to learn the average behaviour over histories $x_t$ with the same value of $\xD_t$.  If $\xD_t$ hides goal-directed information, the default policy will learn behaviour that is generally useful regardless of the current goal. We can make this precise by noting that optimizing the objective \eqref{eq:objective:standard} with respect to $\pp$ amounts to supervised learning of $\pp$ on trajectories generated by $\traj$, i.e.\ this is a distillation process from $\traj$ to $\pp$ \citep{Hinton-distillation, rusu2016policy_distillation,ParisottoBS16,teh2017distral}. In the nonparametric case, the optimal default policy $\pp_*$ can be derived as:
\begin{align}
    \pp_*(a_t|\xD_t) = \frac{
    \sum_{\tilde{t}} \gamma^{\tilde{t}} \int \left(  \mathbbm{1}(\xD_t=\tilde{x}_{\tilde{t}}^\mathcal{D}) \qq(a_{{t}} | \tilde{x}_{\tilde{t}})\right) \traj(\tilde{x}_{\tilde{t}}) d\tilde{x}_{\tilde{t}} 
    }{
    \sum_{\tilde{t}} \gamma^{\tilde{t}} \int \left(  \mathbbm{1}(\xD_t=\tilde{x}_{\tilde{t}}^\mathcal{D}) \right) \traj(\tilde{x}_{\tilde{t}}) d\tilde{x}_{\tilde{t}} 
    },
\end{align}
where $\traj(\tilde{x}_{\tilde{t}})$ is the probability of seeing history $\tilde{x}_{\tilde{t}}$ at time step $\tilde{t}$ under the policy $\qq$, and the indicator $\mathbbm{1}(\xD_t=\tilde{x}_{\tilde{t}}^\mathcal{D})$ is 1 if the goal-agnostic information of the two histories matches and 0 otherwise.  

It is also worth considering the effect of the objective eq.\ \eqref{eq:objective:infoasym} on the learned policy $\qq$. 
%
Since $\pp$ is learned alongside $\qq$ and not specified in advance,
this objective does not favor any particular behavior a priori. Instead it will encourage a solution in which similar behavior will be executed in different parts of the state space that are similar as determined by $\xD_t$, since the policy $\qq$ is regularized towards the default policy $\pp$. More generally, during optimization of $\qq$ the default policy effectively acts like a shaping reward while the entropy contained in the KL discourages deterministic solutions.

\section{Connection to information bottleneck and variational EM}
\label{sec:Connections}

\subsection{Information bottleneck} 
Reinforcement learning objectives with information theoretic constraints have been considered by multiple authors \citep{tishby2011information,still2012information,tiomkin2017unified}. Such constraints can be motivated by the internal computational limitations of the agent, which limit the rate with which information can be extracted from states (or observations) and translated into actions. Such capacity constraints can be expressed via an information theoretic regularization term that is added to the expected reward.
Specializing to our scenario, where the ``information flow'' to be controlled is between the goal-directed history information $x^G_t$ and action $a_t$ (so that the agent prefers default, goal-agnostic, behaviour), consider the objective:  
\begin{align}
\textstyle
    \mathcal{L}_I = \EE_{\traj}\left [ \sum_t  \gamma^t r(s_t,a_t) - \alpha \gamma^t \MutI [ \xG_t, a_t|\xD_t ]  \right],
    \label{eq:objective:IB}
\end{align}
where $\MutI[ \xG_t, a_t|\xD_t ]$ is the conditional mutual information between $\xG_t$ and $a_t$ given $\xD_t$.  The conditional mutual information can be upper bounded:
\begin{align}
    \EE_{\traj}[\MutI [ \xG_t, a_t|\xD_t ]]
    = \EE_{\traj}\left[
    \log\frac{
    \traj(\xG_t|\xD_t)\qq(a_t|\xG_t,\xD_t)
    }{
    \traj(\xG_t|\xD_t)\traj(a_t|\xD_t)
    }\right]
    \le \EE_{\traj}\left[
    \log\frac{
    \qq(a_t|\xG_t,\xD_t)
    }{
    \pp(a_t|\xD_t)
    } \right]
\end{align}
where the inequality is from the fact that the KL divergence  $\KL[\traj(a_t|\xD_t)\|\pp(a_t|\xD_t)]$ is positive \citep[see][]{alemi2016deep}.
Re-introducing this into (\ref{eq:objective:IB}) we find that the KL regularized objective in eq.\ (\ref{eq:objective:infoasym}) can be seen as a lower bound to eq.\ \eqref{eq:objective:IB}, where the agent has a capacity constraint on the channel between goal-directed history information and (future) actions.
%
%
See section \ref{sec:appendix:IB} in the appendix for a generalization including latent variables.
In this light, we can see our work as a particular implementation of the information bottleneck principle, where we penalize the dependence on the information that is hidden from the default policy.




\subsection{Variational EM}

The above setup also bears significant similarity to the training of variational autoencoders \citep{kingma2013auto,rezende14} and, more generally the variational EM framework for learning latent variable models \citep{dempster1977maximum,neal1999learning}. The setup is as follows. Given observations $\mathcal{X} = \{ x_1, \dots x_N \}$ the goal is to maximize the log marginal likelihood $\log p_\theta(\mathcal{X}) = \sum_i \log p_\theta(x_i) $ where $p_\theta(x) = \int p_\theta(x,z) dz$. This marginal likelihood can be bounded from below by $\sum_i \EE_{q_\phi(z|x_i)} [ \log p_\theta(x_i| z) - \log \frac{q_\phi(z | x_i)}{p_\theta(z)}  ]$ with $q_\phi(z|x_i)$ being a learned approximation to the true posterior $p_\theta(z | x_i)$. 
This lower bound exhibits a similar information asymmetry between $q$ and $p$ as the one introduced between $\qq$ and $\pp$ in the objective in eq.\ (\ref{eq:objective:infoasym}). In particular, in the multi-task case  discussed in section \ref{sec:LearnedPriors} with one task per episode, $x_i$ can be seen to take the role of the task, $\log p(x_i | z)$ that of the task reward, $q(z|x_i)$ that of task conditional policy, and $p(z)$ the default policy. Therefore maximizing eq.\ (\ref{eq:objective:infoasym}) can then be thought of as learning a generative model of behaviors that can explain the solution to different tasks.

\section{Algorithm}
\label{sec:Algorithm}

In practice the objective in eq.\ \ref{eq:objective:infoasym} can be optimized in different ways. A simple approach is to perform alternating gradient ascent in $\pp$ and $\qq$. Optimizing $\mathcal{L}$ with respect to $\pp$ amounts to supervised learning with $\qq$ as the data distribution (distilling $\qq$ into $\pp$). Optimizing $\qq$ given $\pp$ requires solving a regularized expected reward problem which can be achieved with a variety of algorithms \citep{schulman2017equivalence,teh2017distral,haarnoja2017reinforcement,hausman2018learning,haarnoja2018soft}.

The specific algorithm choice in our experiments depends on the type of environment. For the continuous control domains we use SVG(0) \citep{heess2015learning} with experience replay and a modification for the KL regularized setting \citep{hausman2018learning,haarnoja2018soft}. The SVG(0) algorithm learns stochastic policies by backpropagation from the action-value function. We estimate the action value function using $K$-step returns and the Retrace operator for low-variance off-policy correction (see \cite{MunosSHB16}; as well as \cite{hausman2018learning,riedmiller2018learning}). For discrete action spaces we use a batched actor-critic algorithm (see \cite{espeholt2018impala}). The algorithm employs a learned state-value function and obtains value estimates for updating the value function and advantages for computing the policy gradient using $K$-step returns in combination with the V-trace operator for off-policy correction. All algorithms are implemented in batched distributed fashion with a single learner and multiple actors.
In algorithm \ref{alg:KStep} we provide pseudo-code for actor-critic version of the algorithm with $K$-step returns. Details of the off-policy versions of the algorithms for continuous and discrete action spaces can be found in the appendix (section \ref{sec:appendix:algorithm}).

\section{Related work}
\label{sec:Related}

There are several well established connections between certain formulations of the reinforcement learning literature and concepts from the probabilistic modeling literature. The formalisms are often closely related although derived from different intuitions, and with different intentions.

\emph{Maximum entropy reinforcement learning}, stochastic optimal control, and related approaches build on the observation that some formulation of the reinforcement learning problem can be interpreted as exact or approximate variational inference in a probabilistic graphical model in which the reward function takes the role of log-likelihood \cite[e.g.][]{ziebart2010modeling,kappen2012optimal,toussaint2009robot}. While the exact formulation and algorithms vary, they result in an entropy or KL regularized expected reward objective. These algorithms were originally situated primarily in the robotics and control literature but there has been a recent surge in interest in deep reinforcement learning community \citep[e.g.][]{fox2015taming,schulman2017equivalence,nachum2017bridging,haarnoja2017reinforcement,hausman2018learning,haarnoja2018soft}.

\begin{algorithm}[t]
\caption{Simple actor-critic algorithm with K-step returns}
\label{alg:KStep}
\begin{algorithmic}
    \State policy: $\qq_\theta$, initial parameters $\theta^{0}$
    \State default policy:  $\pp_\phi$; initial parameters $\phi^{0}$
    \State Q-function:  $Q_\psi$; initial parameters $\psi^{0}$
    \For{j=1, \dots}
        \For{t = 0, K, 2K, \dots T}
            \State rollout partial trajectory: $\tau_{t:t+K} = (s_t, a_t, r_t \dots r_{t+K}) $
            \State compute KL: $\widehat{\KL}_{t'} = \KL[ \qq(\cdot | s_{t'}) \| \pp(\cdot | s_{t'}) ]$
            \State Estimate boostrap value: $\hat{V} = \EE_{\qq(\cdot|s_{t+K})}[ Q(s_{t+K}, a) ] - \alpha \widehat{\KL}_{t+K}$
            \State Estimate Q targets: $\hat{Q}_{t'} = \sum_{t''=t'}^{t+K-1} (r_{t''} - \alpha \widehat{\KL}_{t''}) + \hat{V}$
            \State Agent policy loss: $\hat{L}_\qq = \sum_{t'=t}^{t+K-1} \EE_{\qq(\cdot| s_{t'})} [ Q(s_{t'}, a) ] - \alpha \widehat{\KL}_{t'}$
            \State Q-value loss: $\hat{L}_Q  = \sum_{t'=t}^{t+K-1} \| \hat{Q}_{t'} - Q(s_{t'}, a_{t'}) \|^2$
            \State Default policy loss: $\hat{L}_{\pp} = \sum_{t'=t}^{t+K-1} \widehat{\KL}_{t'}$
            \State $\theta \leftarrow \theta + \beta_{\qq} \nabla_\theta \hat{L}_{\qq}$
            \hspace{0.5cm} $\phi \leftarrow \phi - \beta_{\pp} \nabla_{\phi} \hat{L}_{\pp}$
            \hspace{0.5cm} $\psi \leftarrow \psi - \beta_{Q} \nabla_{\psi} \hat{L}_{Q}$
        \EndFor
    \EndFor
\end{algorithmic}
\end{algorithm}

Related but often seen as distinct is the familiy of \emph{expectation maximization policy search algorithms} \citep[e.g.][]{Peters10,rawlik12,levine2013variational,montgomery2016guided,chebotar2016path,abdolmaleki2018maximum}. These cast policy search as an alternating optimization problem similar to the EM algorithm for learning probabilistic models. They differ in the specific implementation of the equivalents of the E and M steps; intuitively the default policy is repeatedly replaced by a new version of the policy. 

The DISTRAL algorithm \citep{teh2017distral} as well as the present paper can be seen as taking an intermediate position: unlike in the class of RL-as-inference algorithms the default policy is not fixed but learned, but unlike in the classical EM policy search the final result of the optimization remains regularized since the default policy is constrained relative to the policy. As explained above this can be seen as analogous to the relative roles of learned model and observation specific posterior in fitting a generative model. Similar to DISTRAL, Divide and Conquer \citep{ghosh2018divideandconquer} learns an ensemble of policies, each specializing to a particular context, which are regularized towards one another via a symmetric KL penalty, with the behavior of the ensemble distilled to a single fixed policy. In concurrent work~\cite{goyal2019infobot} propose an information bottleneck architecture for policies with latent variables that leads to a KL-regularized formulation similar to the one described in Appendix~\ref{sec:appendix:IB:2}. The information bottleneck is implemented in latent space and the default policy is obtained by marginalization with a goal-agnostic prior.


An important feature of EM policy search and other policy gradient algorithms is the presence of a KL constraint that limits the relative change of the policy to some older version across iterations to control for the rate of change in the policy \cite[e.g.][]{schulman2015trust,heess2015learning,schulman2017ppo,heess2017emergence,nachum2017trust}. The constraint can be implemented in different ways, and collectively the algorithms are often classified as ``trust region'' methods. 
Note that for a KL regularized objective to be a trust region \citep{NoceWrig06}, additional assumptions need to hold. In principle, as an optimization technique, the critical points of the KL regularized objective for some function $f(\theta)$ have to be, provably, the same as for the non-regularized objective. This is not trivial to show unless the trust region for step $k$ is around $\theta_k$. In our case, there is no such guarantee even if we remove the asymmetry in information between default policy and policy or make the default policy be an old copy of the policy.

Other related works motivated from an optimization perspective include Deep Mutual Learning \citep{DML18} applied in supervised learning, where KL-regularization is used with a learned prior that receives the same amount of information as the trained model.
\citet{kirkpatrick2017overcoming} introduces EWC to address catastrophic forgetting, where a second order Taylor expansion of the KL, in a KL-regularized objective, forces the main policy to stay close to solutions of previously encountered tasks. \citet{CzarneckiJJHTHO18} also relies on a KL-regularized objective to ensure policies explored in a curriculum stay close to each other.

Conceptually distinct but formally closely related to maximum entropy and KL-regularized formulations are computational models of bounded rationality \cite[e.g.][]{tishby2011information,ortega2011information,still2012information,rubin2012trading,tiomkin2017unified} which introduce information constraints to account for the agent's internal computational constraints on its ability to process information. As discussed in section \ref{sec:Connections} the present formulation can be seen as a more general formulation of the idea.

\section{Continuous control experiments}
\label{sec:experiments:cc}


In our experiments, we study the effect of using a learned default policy to regularize the behavior of our agents, across a wide range of environments spanning sparse and dense reward tasks. In particular, we evaluate the impact of conditioning the default policy on various information sets $\xD$ on the learning dynamics, and evaluate the potential of pretrained default policies for transfer learning. In these experiments, we consider two streams of information which are fed to our agents: \textbf{task specific information} (task) and \textbf{proprioception} (proprio), corresponding to \textbf{walker} (body) specific observations (joint angles etc.). 
\begin{figure}[!htp]
    \begin{center}
        \subfigure[]{\label{fig:tasks:goto_k_targets}\includegraphics[scale=0.08]{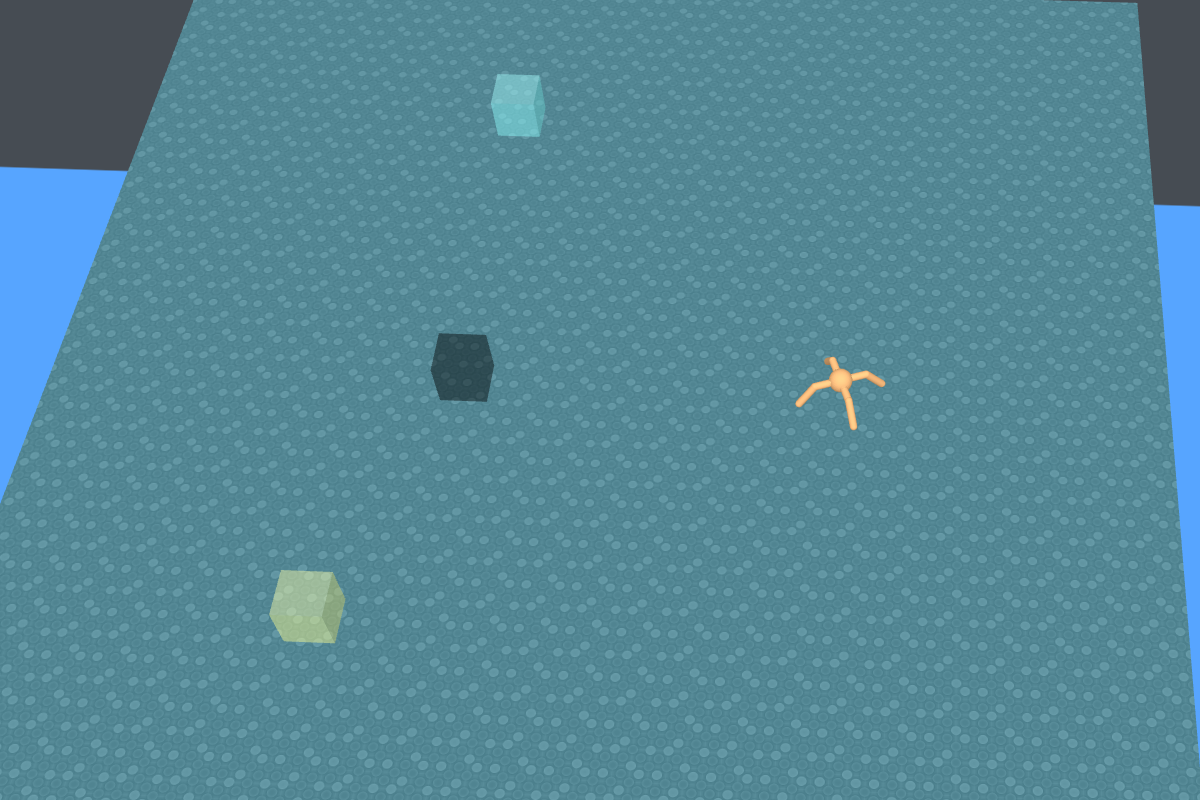}}
        \subfigure[]{\label{fig:tasks:move_box_to_k_targets}\includegraphics[scale=0.08]{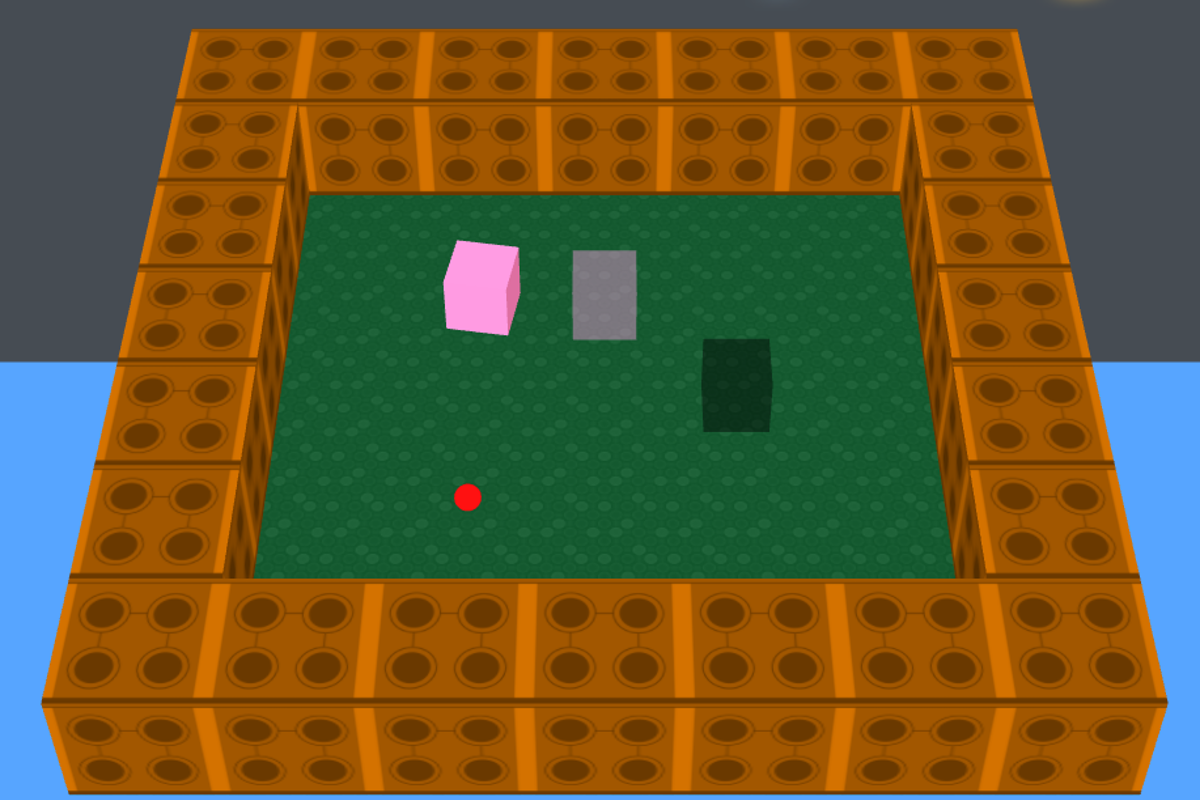}}
        \subfigure[]{\label{fig:tasks:foraging}\includegraphics[scale=0.08]{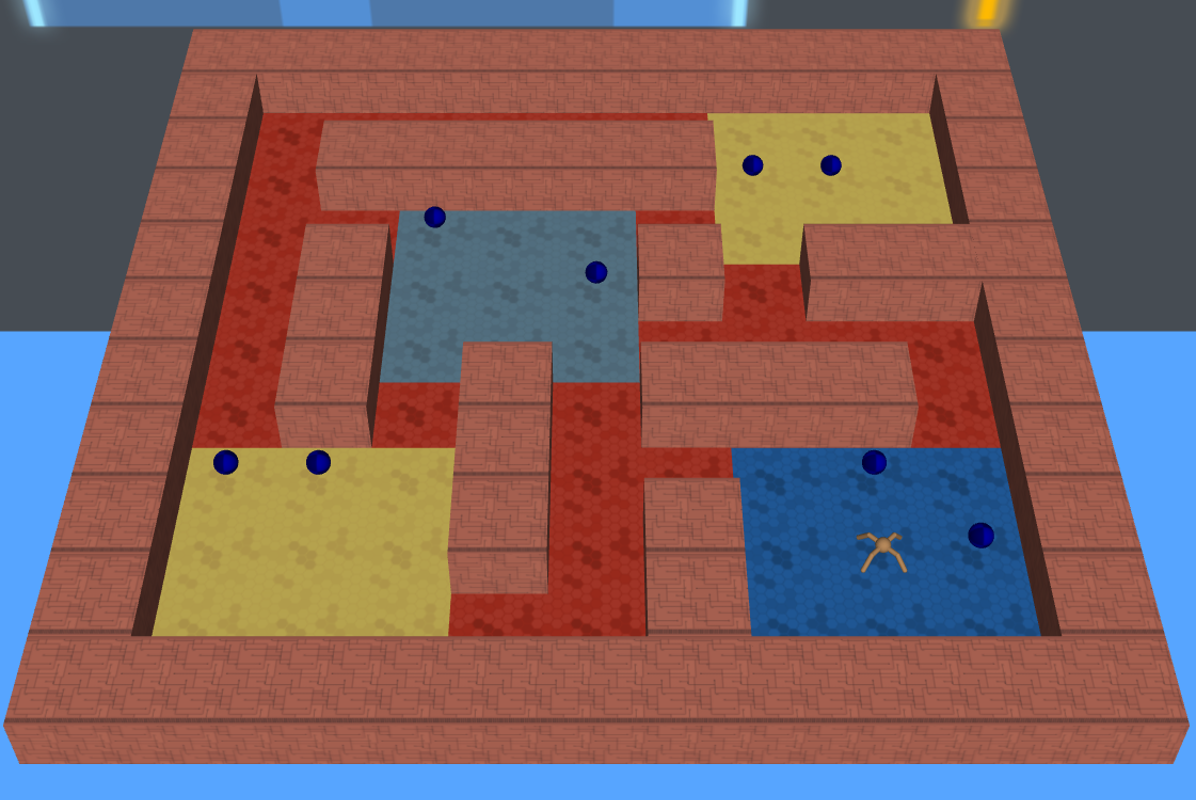}}
        \subfigure[]{\label{fig:tasks:walls}\includegraphics[scale=0.08]{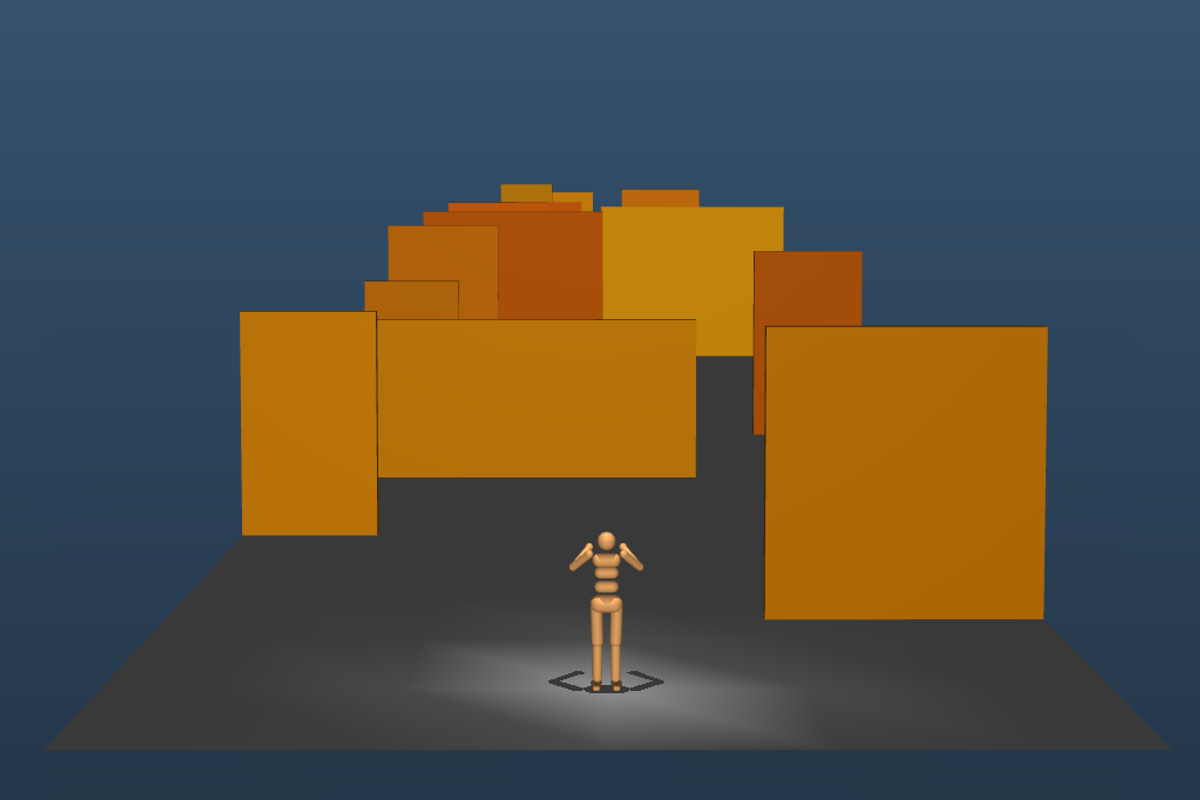}}
        \caption{\textbf{Tasks visualization}. (a): Go to one of K target tasks, with \textit{quadruped}; (b): Move one box to one of K targets task, with \textit{jumping ball} (red); (c): Foraging in the maze task, with \textit{quadruped}; (d): Walls task with \textit{humanoid}, where the goal is avoid walls while running through a terrain.
        }
        \label{fig:tasks_visualization}
    \end{center}
\end{figure}

We consider three walkers: \textit{jumping ball} with 3 degrees of freedom (DoF) and 3 actuators; \textit{quadruped} with 12 DoF and 8 actuators; \textit{humanoid} with 28 DoF and 21 actuators. The task is specified to the agent either via an additional feature vector (referred to as \textit{feature-tasks}) or in the form of visual input (\textit{vision-task}).
%
The tasks differ in the type of reward: in \textit{sparse} reward tasks a non-zero reward is only given when a (sub-)goal is achieved (e.g.\ the  target was reached); in \textit{dense} reward tasks smoothly varying shaping reward is provided (e.g.\ negative distance to the target). We consider the following tasks.

\textbf{Walking task}, a \textit{dense-reward} task based on \textit{features}. The walker needs to move in one of four randomly sampled directions, with a fixed speed; the direction being resampled half-way through the episode. \textbf{Walls task}, a \textit{dense-reward} \textit{vision-task}. Here the walker has to traverse a corridor 
while avoiding walls. \textbf{Go to one of K targets task}, a \textit{sparse-reward} \textit{feature-based} task. The walker has to go to one of K randomly sampled targets. For K=1, the target can either reappear within the episode (referred to as the \textit{moving target} task) or the episode can end upon reaching the target. \textbf{Move one box to one of K targets}, a \textit{sparse-reward} \textit{feature-based-task}. The walker has to move a box to one of K targets, and optionally, go on to one of the remaining targets. The latter is referred to as the \textit{move one box to one of K targets and go to another target}). \textbf{Foraging in the maze task}, a \textit{sparse-reward} \textit{vision-task}. The walker collects apples in a maze. Figure \ref{fig:tasks_visualization} shows visualizations of the walkers and some of the tasks. Refer to appendix~\ref{sec:appendix:walkers_tasks} for more details.



\paragraph{Experimental Setup} As baseline, we consider policies trained with standard entropy regularization. When considering the full training objective of eq.~\ref{eq:objective:standard}, the default policy network shares the same structure as the agent's policy. In both cases, hyper-parameters are optimized on a per-task basis. We employ a distributed actor-learner architecture \citep{espeholt2018impala}: actors execute recent copies of the policy and send data to a replay buffer of fixed size; while the learner samples short trajectory windows from the replay and computes updates to the policy, value, and default policy. We experimented with a number of actors in $\{32, 64, 128, 256\}$ (depending on the task) and a single learner. Results with a single actor are presented in appendix~\ref{sec:appendix:distributed}. Unless otherwise mentioned, we plot average episodic return as a function of the number of environment transitions processed by the learner\footnote{Note that due to the distributed setup with experience replay this number does not directly translate to the number of environment steps executed or gradient updates performed (the latter can be computed dividing the steps processed by batch size and unroll length).}. Each experiment is run with five random seeds. For more details, see appendix~\ref{sec:appendix:algorithm:continuous}



We consider three information sets passed to the default policy: \textbf{proprioceptive}, receiving only proprioceptive information; \textbf{task-subset}, receiving proprioceptive and a subset of task-specific information; \textbf{full-information}, receiving the same information as the policy.

\begin{figure}[!htp]
    \begin{center}
        \includegraphics[scale=0.115]{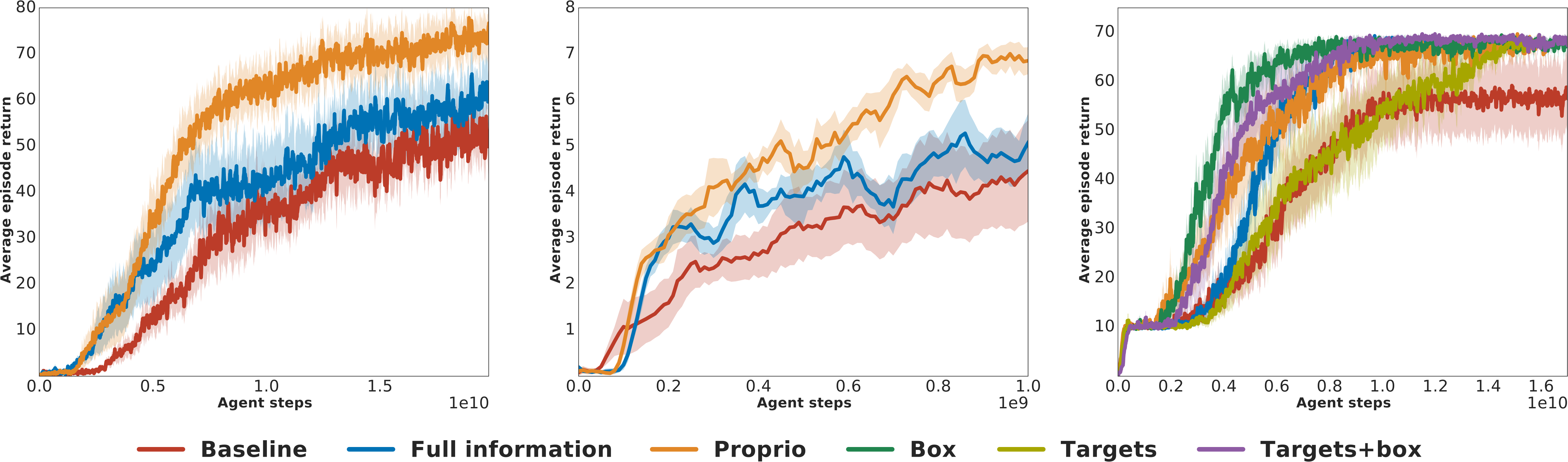}
        \caption{\textbf{Results for the \textit{sparse-reward} tasks with \textit{complex walkers}}. \textbf{Left}: go to moving target task with \textit{humanoid}. \textbf{Center}: foraging in the maze results with \textit{quadruped}. \textbf{Right}: moving one box to one of two targets and go to another target task with \textit{quadruped}. The legends denote additional to the proprioception, information passed to the default policy (except baseline, where we do not use default policy). 
        }
        \label{fig:main_positive_results}
    \end{center}
\end{figure}

The main finding of our experiments is that the default policy with limited task information provides considerable speed-up in terms of learner steps for the \textit{sparse-reward} tasks with \textit{complex walkers} (\textit{quadruped}, \textit{humanoid}). The results on these tasks are presented in figure~\ref{fig:main_positive_results}. More cases are covered in the appendix~\ref{sec:appendix:additional_results}.


Overall, the proprioceptive default policy is very effective and gives the biggest gains in the majority of tasks. Providing additional information to the default policy, leads to an improvement only in a small number of cases (figure~\ref{fig:main_positive_results}, right and appendix~\ref{sec:appendix:additional_results:sparse_reward:ant}). In these cases, the additional information (e.g. box position), provides useful inductive bias for policy learning. For the \textit{dense-reward} tasks or for a simple walker body adding the default policy has limited or no effect (see appendix~\ref{sec:appendix:additional_results:dense_reward},~\ref{sec:appendix:additional_results:sparse_reward:bb8}). We hypothesize that this is due to the relative simplicity of the regular policy learning versus the KL-regularized setup.
In the case of \textit{dense-reward} tasks the agent receives a strong learning signal. For simple walkers there may be little behavioral structure that can be modeled by the default policy. Finally, when the default policy receives full information, the optimal default policy could, in principle, exactly copy the agent policy, thus bearing similarity to methods which regularize the current policy against an older copy of itself. In all cases, the default policy is likely to be less effective at encouraging the generalization of behavior across different contexts and thus to provide less meaningful regularization.

\begin{figure}[!htp]
    \begin{center}
        \includegraphics[scale=0.115]{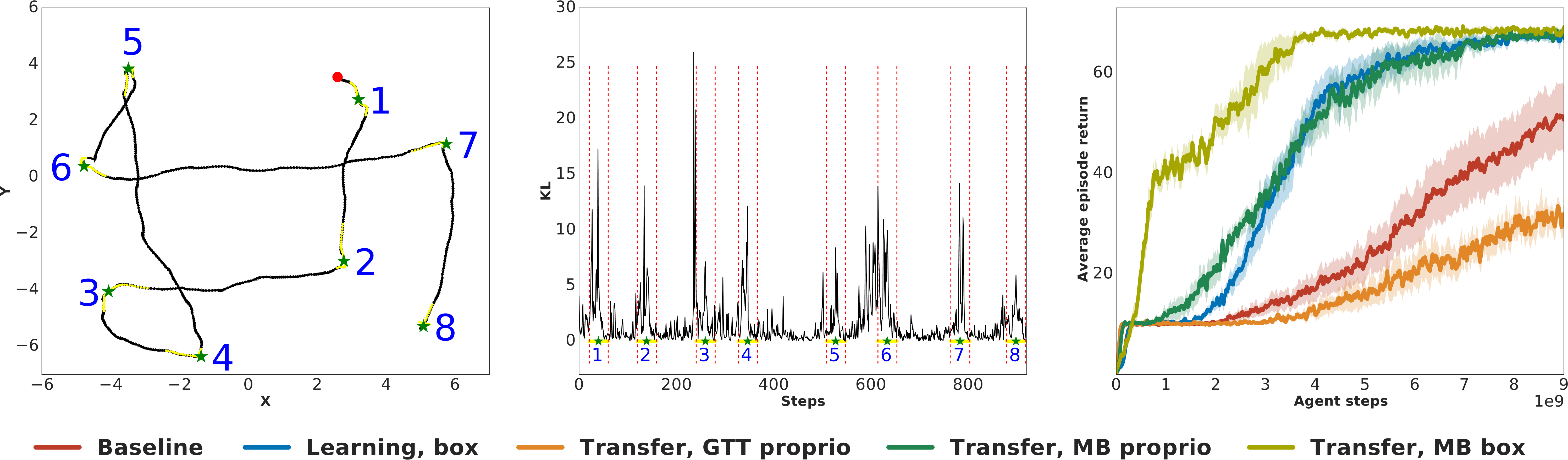}
        \caption{\textbf{Behavior analysis and transfer results}. \textbf{Left}: the trajectory of the agent on go to moving target task with \textit{quadruped}. \textbf{Center}: KL divergence from the agent policy to the proprioceptive default policy plotted over time for the same trajectory. \textbf{Right}: Performance of the transfer on move one box to one of 3 targets task with \textit{quadruped}. The legend whether the default policy is learned or is transferred. Furthermore, it specifies the task from which the default policy is transferred as well as additional information other than the proprioceptive information that the default policy is conditioned on, if any.}
        \label{fig:goto_qualitative}
    \end{center}
\end{figure}    

We analyze the agent behavior on the go to moving target task with a \textit{quadruped} walker. We illustrate the agent trajectory for this task in figure~\ref{fig:goto_qualitative}, left. The red dot corresponds to the agent starting position. The green stars on the left and central figures correspond to the locations of the targets with blue numbers indicating the order of achieving the targets. The yellow dots on the left and central curves indicate the segment (of 40 time steps) near the target. In figure~\ref{fig:goto_qualitative}, center, we show the KL divergence, $KL[\qq \| \pp]$, from the agent policy to the proprioceptive default policy. We observe that for the segments which are close to the target (yellow dots near green star), the value of the KL divergence is high. In these segments the walker has to stop and turn in order to go to another target. It represents a deviation from the standard, walking behavior, and we can observe it as spikes in the KL. Furthermore, for the segments between the targets, e.g. 4 --> 5, the KL is much lower. 

\paragraph{Default Policy Transfer} We additionally explore the possibility of reusing pretrained default policies to regularize learning on new tasks. Our transfer task is \textit{moving one box to one of 2 targets and going to another target task} with the \textit{quadruped}. We consider different default policies: \textbf{GTT proprio}: proprioceptive information only trained on \textit{going to moving target task} (\textit{GTT}); \textbf{MB proprio}: proprioceptive information only trained on \textit{moving one box to one target task} (\textit{MB}); \textbf{MB box}: similar \emph{MB proprio}, but with box position information as additional input. The results are given in figure~\ref{fig:goto_qualitative}, right. 
We observe a significant improvement in learning speed transferring the pretrained default policies to the new task. Performance improves as the trajectory distribution modeled by the default policy is closer to the one appropriate for the transfer task (compare GTT proprio with MB proprio; and MB proprio with MB box).

\begin{figure}[!htp]
    \begin{center}
        \includegraphics[scale=0.09]{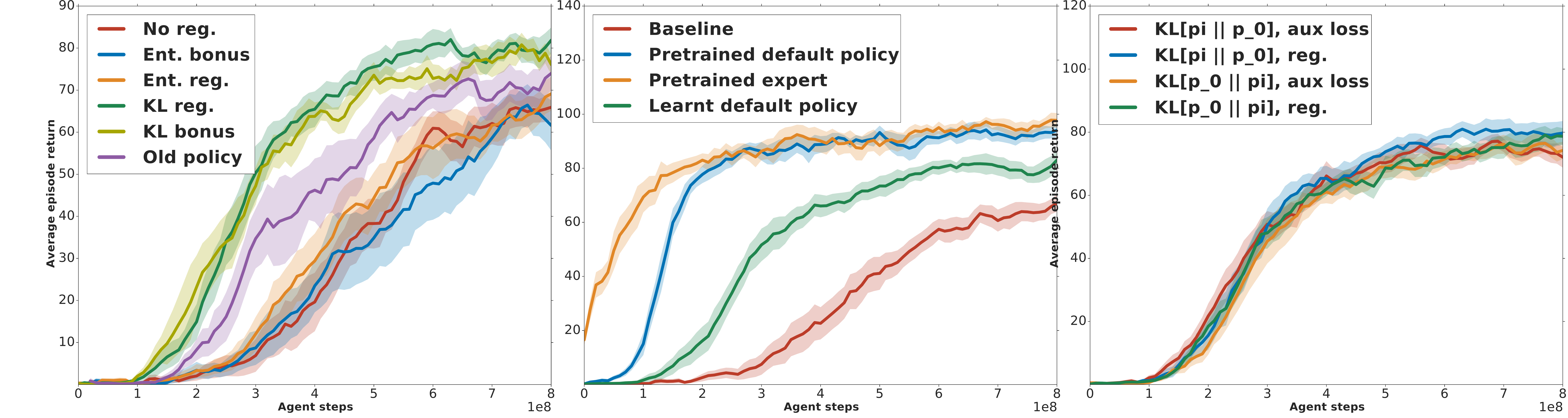}
        \caption{\textbf{Ablations} on \textit{go to moving target} task with \textit{quadruped} body. \textbf{Left}: Comparing various regularization schemes.
        \textbf{Center}: Optimistic baselines comparing pretrained default policies.
        \textbf{Right}: Analysis of the order of the default policy in the KL-term.
        }
        \label{fig:ablations}
    \end{center}
\end{figure}




\paragraph{Ablative Analysis} To gain deeper insights into our method, we compare different forms of regularization of the standard RL objective: \emph{entropy bonus} - adding an entropy term $H(\pi(\cdot | s_{t}))$ to the per-timestep actor loss; \emph{entropy regularization} - optimizing the objective~\eqref{eq:objective:entropy}; \emph{KL bonus} - adding the KL-divergence term $\KL \left [ \qq(a_t|s_t) \| \pp(a_t | s_t) \right ]$ from the agent policy to the default one to the per-timestep actor loss; \emph{KL-regularization} - optimizing the objective~\eqref{eq:objective:standard}; \emph{KL-regularization to the old policy} - optimization of the objective~\ref{eq:objective:standard} where regularization is done wrt. an older version of the main policy (updated every 100 steps). The default policy receives only proprioceptive information in these experiments. The task is \textit{go to moving target}. 
As can be seen in Figure~\ref{fig:ablations} left, all three KL-based variants improve performance over the baseline, but regularizing against the information restricted default policy outperforms regularization against an old version of the policy.

Furthermore, we assess the benefit of the KL-regularized objective~\ref{eq:objective:standard} when used with an idealized default policy. We repeat the go-to-target experiment with a pretrained default policy on the same task. Figure~\ref{fig:ablations} center, shows a significant difference between the baseline and different regularization variants: using the pretrained default policy, learning the default policy alongside the main policy or using a pretrained \emph{expert} (default policy with access to the full state). This suggests that large gains may be achievable in situations when a good default policy is known \emph{a priori}. We performed the same analysis for the dense reward but we did not notice any gain. The speed-up from regularizing to the pretrained \emph{expert} is significant, however it  corresponds to regularizing against an existing solution and can thus primarily be used as a method to speed-up the experiment cycles, as it was demonstrated in the kickstarting framework~\citep{schmitt2018kickstarting}.


Finally, we study impact of the direction of the KL in  objective~\ref{eq:objective:standard} on the learning dynamics. Motivated by the work on policy distillation~\citep{rusu2016policy_distillation} we flip the KL and use $\KL \left [ \pp(a_t|s_t) \| \qq(a_t | s_t) \right ] $ instead of the described before $\KL \left [ \qq(a_t|s_t) \| \pp(a_t | s_t) \right ]$. We use this term either in per time step actor loss (auxiliary loss) or as a regularizer The figure~\ref{fig:ablations} right, shows that there is no significant difference between these regularization schemes, which suggests that the idea of learned default policy can be viewed from student-teacher perspective, where default policy plays the role of the teacher. This teacher can be used in a new task.

\section{Discrete action spaces experiments}


We also evaluate our method on the DMLab-30  set of environments. DMLab \citep{beattie2016dmlab} provides a suite of rich, first-person environments with tasks ranging from complex navigation and laser-tag to language-instructed goal finding. Recent works on multitask training~\citep{espeholt2018impala} in this domain have used a form of batched-A2C with the V-trace algorithm to maximize an approximation of the entropy regularized objective described earlier, where the default policy is a uniform distribution over the actions. 

Typically, the agent receives visual information at each step, along with an instruction channel used in a subset of tasks. The agent receives no task identifier. We adopt the architecture employed in previous work~\citep{espeholt2018impala} in which frames, past actions and rewards are passed successively through a deep residual network and LSTM, finally predicting a policy and value function. All our experiments are tuned with population-based training~\citep{jaderberg2017pbt}. Further details are provided in appendix~\ref{sec:appendix:algorithm:discrete}.

DMLab exposes a large action space, specifically the cartesian product of atomic actions along seven axes. However, commonly a human-engineered restricted subset of these actions is used at training and test time, simplifying the exploration problem for the agent. For example, the used action space has a \emph{forward bias}, with more actions resulting in the agent moving forward rather than backwards. This helps with exploration in navigation tasks, where even a random walk can get the agent to move away from the starting position.
The uniform default policy is used on top of this human engineered small action space, where its semantics are clear. 

In this work, we instead consider a much larger combinatorial space of actions. We show that a pure uniform default policy is in fact unhelpful when human knowledge is removed from defining the right subset of actions to be uniform over, and the agent under-performs. Learning the default policy, even in the extreme case when the default policy is not conditioned on any state information, helps recovering which actions are worth exploring and leads to the emergence of a useful action space without any hand engineering.



\begin{figure}[!htp]
    \begin{center}
      \subfigure{ \includegraphics[width=0.3\textwidth]{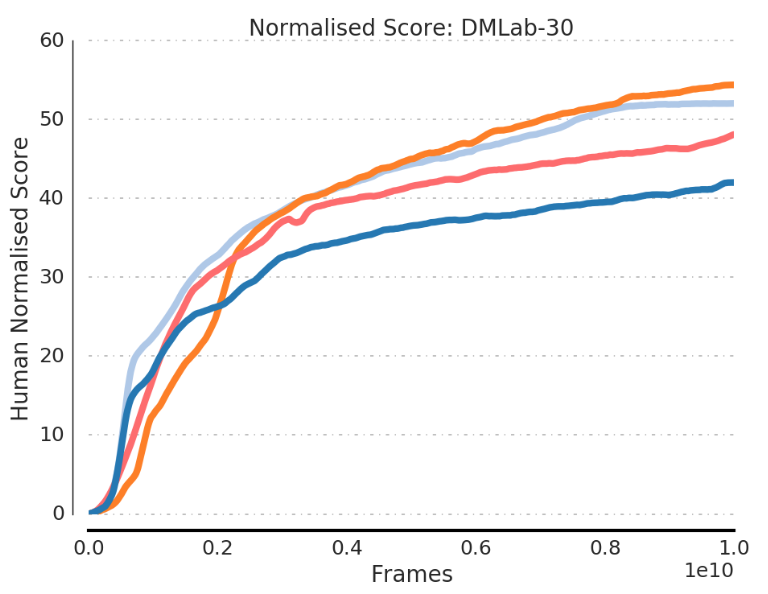}
        \includegraphics[width=0.3\textwidth]{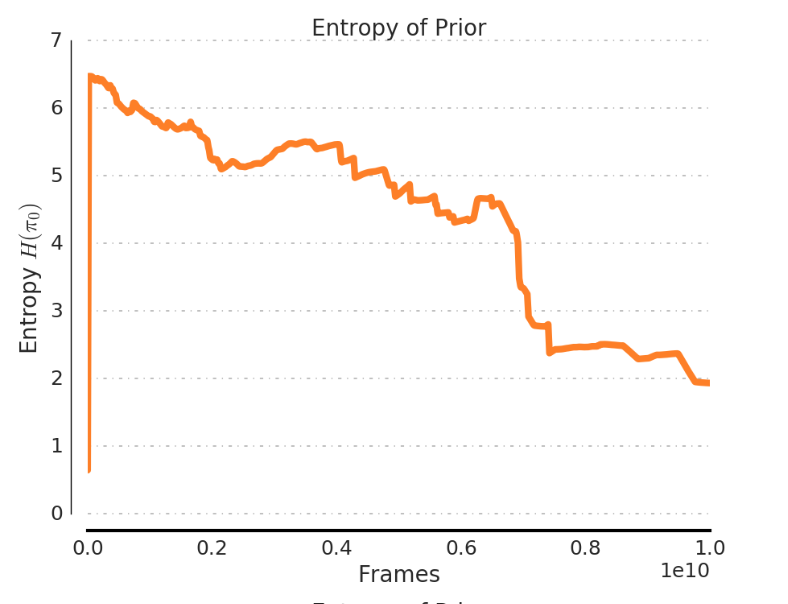}
       
      \includegraphics[width=0.4\textwidth]{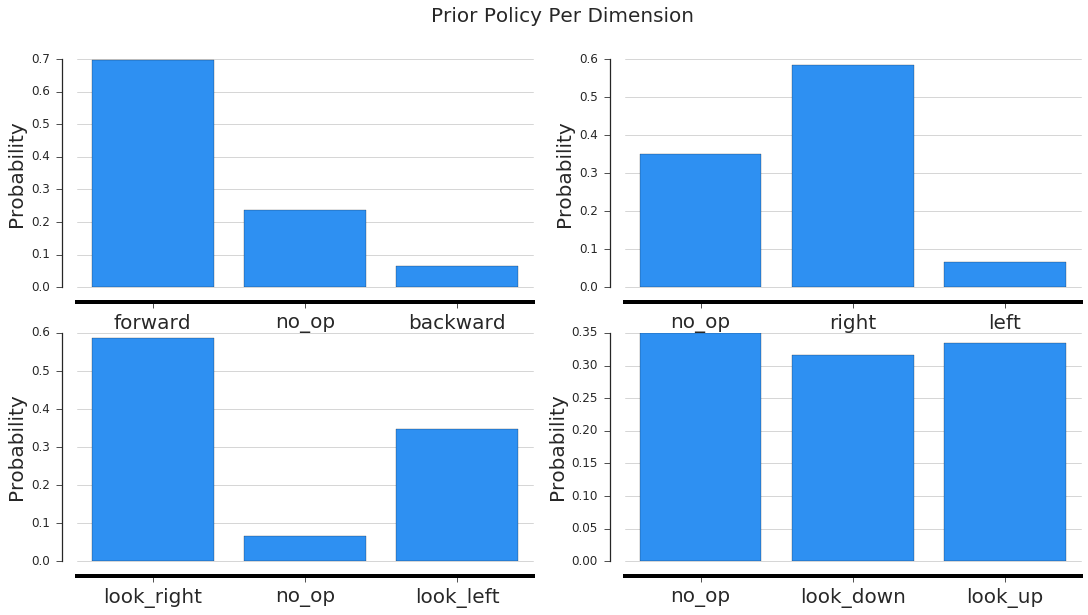}}
    
         \includegraphics[width=1.0\textwidth]{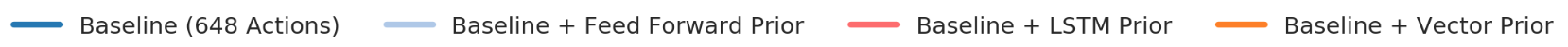}
        \caption{\textbf{DMLab30}. Left, comparison between baseline (same as \cite{espeholt2018impala}) that uses uniform distribution over actions as a default policy and three different possible default policies. Center, the entropy for the vector default policy over learning. Right, marginalized distribution over few actions of interest for the vector default policy. }
        \label{fig:dmlab}
    \end{center}
\end{figure}

Figure~\ref{fig:dmlab} shows the results of our experiments. We consider a flat action space of 648 actions, each moving the agent in different spatial dimensions. We run the agent from \citep{espeholt2018impala} as baseline which is equivalent to considering the default policy to be a uniform distribution over the 648 actions, and three variants of our approach, where the default policy is actually learnt. 

For \textbf{feed forward default policy}, while the agent is recurrent, the default policy is not. 
That is the policy $\qq$ is conditioned on the full trace of observed states 
$s_1, a_1,.. s_t$,  while the default policy $\pp$ is conditioned only on the current frame $a_{t-1}, s_t$. Given that most of the 30 tasks considered require memory in order to be solvable, the default policy has to generalize over important task details. 
\textbf{LSTM default policy} on the other hand, while being recurrent as the agent, it observes only the previous action $a_{t-1}$ and does not receive any other state information. In this instance, the default policy can only model the most likely actions given recent behaviour $a_1, .. a_{t-1}$ in absence of any visual stimuli. For example, if previous actions are \emph{moving forward}, the default policy might predict \emph{moving forward} as the  next action too.  
This is because the agent usually moves consistently in any given direction in order to navigate efficiently.
Finally, the \textbf{vector default policy} refers to a default policy that is independent of actions and states (i.e. average behaviour over all possible histories of states and actions). 

Using any of the default policies outperforms the baseline, with LSTM default policy slightly underperforming compared with the others. The vector default policy performs surprisingly well, highlighting that for DMLab defining a meaningful action space is extremely important for solving the task. Our approach can provide a mechanism for identifying this action space without requiring human expert knowledge on the tasks. Note in middle plot, figure~\ref{fig:dmlab}, that the entropy of the default policy over learning frames goes down, indicating that the default policy becomes peaky and is quite different from the uniform distribution which the baseline assumes. Note that when running the same experiments with the original human-engineered smaller action space, no gains are observed. This is similar to the continuous control setup, corresponding to changing the walker to a simple one and hence converting the task into a denser reward one. 

Additionally, in figure~\ref{fig:dmlab} right, for the vector default policy, we show the probability of a few actions of interest by  marginalizing over all other actions. We notice that the agent has a tendency of moving forward $70\%$, while moving backwards is quite unlikely $10\%$. The default policy discovers one element of the human defined action space, namely \emph{forward-bias} which is quite useful for exploring the map. 
The uniform bias would put same weight for moving forward as for moving backwards, making exploration harder.
We also note that the agent has a tendency to turn right and look right. Given that each episode involves navigating a new sampled map, such a bias provides a meaningful exploration boost, as it suggest a \emph{following the wall} strategy, where at any new intersection the agent 
always picks the same turning direction (e.g. right) to avoid moving in circles. But as expected, since neither looking up or looking down provides any advantage, these actions are equally 
probable.

\section{Discussion and Conclusions}

In this work we studied the influence of learning the default policy in the KL-regularized RL objective. Specifically we looked at the scenario where we enforce information asymmetry between the default policy and the main one. In the continuous control, we showed empirically that in the case of \textit{sparse-reward} tasks with complex walkers, there is a significant speed-up of learning compared to the baseline. In addition, we found that the gain in \textit{dense-reward} tasks and/or with simple walkers was limited, or difficult to get.
Moreover, we demonstrated that significant gains can be achieved in the discrete action spaces. We provided evidence that these gains are mostly due to the information asymmetry between the agent and the default policy.
Best results are obtained when the default policy sees only a subset of information, allowing it to learn task-agnostic behaviour. Furthermore, these default polices can be reused to significantly 
speed-up learning on new tasks.  

\section{Acknowledgments}
    The authors would like to thank Abbas Abdolmaleki, Arun Ahuja, Jost Tobias Springenberg, Siqi Liu for their help on experimental side. Furthermore, The authors would like to thank Greg Wayne for useful discussions. Finally, the authors are very grateful to Simon Osindero and Phil Blunsom for their insightful feedback on the paper.

\bibliography{distral}

\begin{thebibliography}{56}
\providecommand{\natexlab}[1]{#1}
\providecommand{\url}[1]{\texttt{#1}}
\expandafter\ifx\csname urlstyle\endcsname\relax
  \providecommand{\doi}[1]{doi: #1}\else
  \providecommand{\doi}{doi: \begingroup \urlstyle{rm}\Url}\fi

\bibitem[Abdolmaleki et~al.(2018)Abdolmaleki, Springenberg, Tassa, Munos,
  Heess, and Riedmiller]{abdolmaleki2018maximum}
Abbas Abdolmaleki, Jost~Tobias Springenberg, Yuval Tassa, R{\'{e}}mi Munos,
  Nicolas Heess, and Martin~A. Riedmiller.
\newblock Maximum a posteriori policy optimisation.
\newblock \emph{CoRR}, abs/1806.06920, 2018.

\bibitem[Alemi et~al.(2016)Alemi, Fischer, Dillon, and Murphy]{alemi2016deep}
Alexander~A. Alemi, Ian Fischer, Joshua~V. Dillon, and Kevin Murphy.
\newblock Deep variational information bottleneck.
\newblock \emph{CoRR}, abs/1612.00410, 2016.
\newblock URL \url{http://arxiv.org/abs/1612.00410}.

\bibitem[Alemi et~al.(2017)Alemi, Poole, Fischer, Dillon, Saurous, and
  Murphy]{alemi2017fixing}
Alexander~A. Alemi, Ben Poole, Ian Fischer, Joshua~V. Dillon, Rif~A. Saurous,
  and Kevin Murphy.
\newblock Fixing a broken elbo.
\newblock \emph{CoRR}, abs/1711.00464, 2017.
\newblock URL \url{http://arxiv.org/abs/1711.00464}.

\bibitem[Beattie et~al.(2016)Beattie, Leibo, Teplyashin, Ward, Wainwright,
  Küttler, Lefrancq, Green, Valdés, Sadik, Schrittwieser, Anderson, York,
  Cant, Cain, Bolton, Gaffney, King, Hassabis, Legg, and
  Petersen]{beattie2016dmlab}
Charles Beattie, Joel~Z. Leibo, Denis Teplyashin, Tom Ward, Marcus Wainwright,
  Heinrich Küttler, Andrew Lefrancq, Simon Green, Víctor Valdés, Amir Sadik,
  Julian Schrittwieser, Keith Anderson, Sarah York, Max Cant, Adam Cain, Adrian
  Bolton, Stephen Gaffney, Helen King, Demis Hassabis, Shane Legg, and Stig
  Petersen.
\newblock Deepmind lab.
\newblock \emph{arXiv preprint arXiv:1612.03801}, 2016.
\newblock URL \url{https://arxiv.org/abs/1612.03801}.

\bibitem[Chebotar et~al.(2016)Chebotar, Kalakrishnan, Yahya, Li, Schaal, and
  Levine]{chebotar2016path}
Yevgen Chebotar, Mrinal Kalakrishnan, Ali Yahya, Adrian Li, Stefan Schaal, and
  Sergey Levine.
\newblock Path integral guided policy search.
\newblock \emph{CoRR}, abs/1610.00529, 2016.

\bibitem[Czarnecki et~al.(2018)Czarnecki, Jayakumar, Jaderberg, Hasenclever,
  Teh, Heess, Osindero, and Pascanu]{CzarneckiJJHTHO18}
Wojciech~Marian Czarnecki, Siddhant~M. Jayakumar, Max Jaderberg, Leonard
  Hasenclever, Yee~Whye Teh, Nicolas Heess, Simon Osindero, and Razvan Pascanu.
\newblock Mix {\&} match agent curricula for reinforcement learning.
\newblock In \emph{{ICML}}, 2018.

\bibitem[Dempster et~al.(1977)Dempster, Laird, and Rubin]{dempster1977maximum}
A.~P. Dempster, N.~M. Laird, and D.~B. Rubin.
\newblock Maximum likelihood from incomplete data via the {EM} algorithm.
\newblock \emph{Journal of the Royal Statistical Society: Series B},
  39:\penalty0 1--38, 1977.

\bibitem[Espeholt et~al.(2018)Espeholt, Soyer, Munos, Simonyan, Mnih, Ward,
  Doron, Firoiu, Harley, Dunning, Legg, and Kavukcuoglu]{espeholt2018impala}
Lasse Espeholt, Hubert Soyer, Remi Munos, Karen Simonyan, Volodymir Mnih, Tom
  Ward, Yotam Doron, Vlad Firoiu, Tim Harley, Iain Dunning, Shane Legg, and
  Koray Kavukcuoglu.
\newblock Scalable distributed deep-rl with importance weighted actor-learner
  architectures.
\newblock \emph{arXiv:1802.01561}, 2018.
\newblock URL \url{https://arxiv.org/abs/1802.01561}.

\bibitem[Fox et~al.(2015)Fox, Pakman, and Tishby]{fox2015taming}
Roy Fox, Ari Pakman, and Naftali Tishby.
\newblock G-learning: Taming the noise in reinforcement learning via soft
  updates.
\newblock \emph{CoRR}, abs/1512.08562, 2015.

\bibitem[Ghosh et~al.(2018)Ghosh, Singh, Rajeswaran, Kumar, and
  Levine]{ghosh2018divideandconquer}
Dibya Ghosh, Avi Singh, Aravind Rajeswaran, Vikash Kumar, and Sergey Levine.
\newblock Divide-and-conquer reinforcement learning.
\newblock In \emph{International Conference on Learning Representations}, 2018.

\bibitem[Goyal et~al.(2019)Goyal, Islam, Strouse, Ahmed, Botvinick, Larochelle,
  Bengio, and Levine]{goyal2019infobot}
Anirudh Goyal, Riashat Islam, Daniel Strouse, Zafarali Ahmed, Matthew
  Botvinick, Hugo Larochelle, Yoshua Bengio, and Sergey Levine.
\newblock Infobot: Transfer and exploration via the information bottleneck.
\newblock \emph{ICLR}, abs/1901.10902, 2019.
\newblock URL \url{http://arxiv.org/abs/1901.10902}.

\bibitem[Haarnoja et~al.(2017)Haarnoja, Tang, Abbeel, and
  Levine]{haarnoja2017reinforcement}
Tuomas Haarnoja, Haoran Tang, Pieter Abbeel, and Sergey Levine.
\newblock Reinforcement learning with deep energy-based policies.
\newblock \emph{CoRR}, abs/1702.08165, 2017.

\bibitem[Haarnoja et~al.(2018)Haarnoja, Zhou, Abbeel, and
  Levine]{haarnoja2018soft}
Tuomas Haarnoja, Aurick Zhou, Pieter Abbeel, and Sergey Levine.
\newblock Soft actor-critic: Off-policy maximum entropy deep reinforcement
  learning with a stochastic actor.
\newblock \emph{CoRR}, abs/1801.01290, 2018.
\newblock URL \url{http://arxiv.org/abs/1801.01290}.

\bibitem[Hausman et~al.(2018)Hausman, Springenberg, Wang, Heess, and
  Riedmiller]{hausman2018learning}
Karol Hausman, Jost~Tobias Springenberg, Ziyu Wang, Nicolas Heess, and Martin
  Riedmiller.
\newblock Learning an embedding space for transferable robot skills.
\newblock \emph{International Conference on Learning Representations}, 2018.
\newblock URL \url{https://openreview.net/forum?id=rk07ZXZRb}.

\bibitem[He et~al.(2015)He, Zhang, Ren, and Sun]{he2015resnet}
Kaiming He, Xiangyu Zhang, Shaoqing Ren, and Jian Sun.
\newblock Deep residual learning for image recognition.
\newblock \emph{arXiv:1512.03385}, 2015.
\newblock URL \url{https://arxiv.org/abs/1512.03385}.

\bibitem[Heess et~al.(2015)Heess, Wayne, Silver, Lillicrap, Erez, and
  Tassa]{heess2015learning}
Nicolas Heess, Gregory Wayne, David Silver, Tim Lillicrap, Tom Erez, and Yuval
  Tassa.
\newblock Learning continuous control policies by stochastic value gradients.
\newblock In \emph{Advances in Neural Information Processing Systems}, pp.\
  2944--2952, 2015.

\bibitem[Heess et~al.(2017)Heess, Sriram, Lemmon, Merel, Wayne, Tassa, Erez,
  Wang, Eslami, Riedmiller, et~al.]{heess2017emergence}
Nicolas Heess, Srinivasan Sriram, Jay Lemmon, Josh Merel, Greg Wayne, Yuval
  Tassa, Tom Erez, Ziyu Wang, Ali Eslami, Martin Riedmiller, et~al.
\newblock Emergence of locomotion behaviours in rich environments.
\newblock \emph{arXiv preprint arXiv:1707.02286}, 2017.

\bibitem[Hinton et~al.(2015)Hinton, Vinyals, and Dean]{Hinton-distillation}
Geoffrey Hinton, Oriol Vinyals, and Jeffrey Dean.
\newblock Distilling the knowledge in a neural network.
\newblock In \emph{NIPS Deep Learning and Representation Learning Workshop},
  2015.
\newblock URL \url{http://arxiv.org/abs/1503.02531}.

\bibitem[Jaderberg et~al.(2017)Jaderberg, Dalibard, Osindero, Czarnecki,
  Donahue, Razavi, Vinyals, Green, Dunning, Simonyan, Fernando, and
  Kavukcuoglu]{jaderberg2017pbt}
Max Jaderberg, Valentin Dalibard, Simon Osindero, Wojciech~M. Czarnecki, Jeff
  Donahue, Ali Razavi, Oriol Vinyals, Tim Green, Iain Dunning, Karen Simonyan,
  Chrisantha Fernando, and Koray Kavukcuoglu.
\newblock Population based training of neural networks.
\newblock \emph{arXiv preprint arXiv:1711.09846}, 2017.
\newblock URL \url{https://arxiv.org/abs/1711.09846}.

\bibitem[Kappen et~al.(2012)Kappen, G{\'o}mez, and Opper]{kappen2012optimal}
Hilbert~J. Kappen, Vicen{\c{c}} G{\'o}mez, and Manfred Opper.
\newblock Optimal control as a graphical model inference problem.
\newblock \emph{Machine Learning}, 87\penalty0 (2):\penalty0 159--182, May
  2012.
\newblock ISSN 1573-0565.

\bibitem[Kingma \& Welling(2013)Kingma and Welling]{kingma2013auto}
Diederik~P Kingma and Max Welling.
\newblock Auto-encoding variational bayes.
\newblock \emph{arXiv preprint arXiv:1312.6114}, 2013.

\bibitem[Kirkpatrick et~al.(2017)Kirkpatrick, Pascanu, Rabinowitz, Veness,
  Desjardins, Rusu, Milan, Quan, Ramalho, Grabska-Barwinska,
  et~al.]{kirkpatrick2017overcoming}
James Kirkpatrick, Razvan Pascanu, Neil Rabinowitz, Joel Veness, Guillaume
  Desjardins, Andrei~A Rusu, Kieran Milan, John Quan, Tiago Ramalho, Agnieszka
  Grabska-Barwinska, et~al.
\newblock Overcoming catastrophic forgetting in neural networks.
\newblock \emph{Proceedings of the National Academy of Sciences}, pp.\
  201611835, 2017.

\bibitem[Kool \& Botvinick(2018)Kool and Botvinick]{wouter2018mental}
Wouter Kool and Matthew Botvinick.
\newblock Mental labour.
\newblock \emph{Nature Human Behaviour}, 2018.
\newblock \doi{10.1038/s41562-018-0401-9}.

\bibitem[Levine \& Koltun(2013)Levine and Koltun]{levine2013variational}
Sergey Levine and Vladlen Koltun.
\newblock Variational policy search via trajectory optimization.
\newblock In \emph{Advances in Neural Information Processing Systems}, pp.\
  207--215, 2013.

\bibitem[Mnih et~al.(2015)Mnih, Kavukcuoglu, Silver, Rusu, Veness, Bellemare,
  Graves, Riedmiller, Fidjeland, Ostrovski, et~al.]{mnih2015human}
Volodymyr Mnih, Koray Kavukcuoglu, David Silver, Andrei~A Rusu, Joel Veness,
  Marc~G Bellemare, Alex Graves, Martin Riedmiller, Andreas~K Fidjeland, Georg
  Ostrovski, et~al.
\newblock Human-level control through deep reinforcement learning.
\newblock \emph{Nature}, 518\penalty0 (7540):\penalty0 529--533, 2015.

\bibitem[Mnih et~al.(2016)Mnih, Badia, Mirza, Graves, Lillicrap, Harley,
  Silver, and Kavukcuoglu]{mnih2016asynchronous}
Volodymyr Mnih, Adria~Puigdomenech Badia, Mehdi Mirza, Alex Graves, Timothy~P
  Lillicrap, Tim Harley, David Silver, and Koray Kavukcuoglu.
\newblock Asynchronous methods for deep reinforcement learning.
\newblock In \emph{International Conference on Machine Learning (ICML)}, 2016.

\bibitem[Montgomery \& Levine(2016)Montgomery and Levine]{montgomery2016guided}
William Montgomery and Sergey Levine.
\newblock Guided policy search as approximate mirror descent.
\newblock \emph{CoRR}, abs/1607.04614, 2016.
\newblock URL \url{http://arxiv.org/abs/1607.04614}.

\bibitem[Munos et~al.(2016)Munos, Stepleton, Harutyunyan, and
  Bellemare]{MunosSHB16}
R{\'{e}}mi Munos, Tom Stepleton, Anna Harutyunyan, and Marc~G. Bellemare.
\newblock Safe and efficient off-policy reinforcement learning.
\newblock In \emph{Advances in Neural Information Processing Systems 29: Annual
  Conference on Neural Information Processing Systems 2016, December 5-10,
  2016, Barcelona, Spain}, pp.\  1046--1054, 2016.
\newblock URL
  \url{http://papers.nips.cc/paper/6538-safe-and-efficient-off-policy-reinforcement-learning}.

\bibitem[Nachum et~al.(2017{\natexlab{a}})Nachum, Norouzi, Xu, and
  Schuurmans]{nachum2017bridging}
Ofir Nachum, Mohammad Norouzi, Kelvin Xu, and Dale Schuurmans.
\newblock Bridging the gap between value and policy based reinforcement
  learning.
\newblock \emph{CoRR}, abs/1702.08892, 2017{\natexlab{a}}.

\bibitem[Nachum et~al.(2017{\natexlab{b}})Nachum, Norouzi, Xu, and
  Schuurmans]{nachum2017trust}
Ofir Nachum, Mohammad Norouzi, Kelvin Xu, and Dale Schuurmans.
\newblock Trust-pcl: An off-policy trust region method for continuous control.
\newblock \emph{CoRR}, abs/1707.01891, 2017{\natexlab{b}}.

\bibitem[Neal \& Hinton(1999)Neal and Hinton]{neal1999learning}
Radford~M. Neal and Geoffrey~E. Hinton.
\newblock Learning in graphical models.
\newblock chapter A View of the EM Algorithm That Justifies Incremental,
  Sparse, and Other Variants, pp.\  355--368. MIT Press, Cambridge, MA, USA,
  1999.

\bibitem[Nocedal \& Wright(2006)Nocedal and Wright]{NoceWrig06}
Jorge Nocedal and Stephen~J. Wright.
\newblock \emph{Numerical Optimization}.
\newblock Springer, New York, NY, USA, second edition, 2006.

\bibitem[Ortega \& Braun(2011)Ortega and Braun]{ortega2011information}
Pedro~A. Ortega and Daniel~A. Braun.
\newblock Information, utility {\&}amp; bounded rationality.
\newblock \emph{CoRR}, abs/1107.5766, 2011.

\bibitem[Ortega \& Braun(2013)Ortega and Braun]{ortega2013thermodynamics}
Pedro~A. Ortega and Daniel~A. Braun.
\newblock Thermodynamics as a theory of decision-making with
  information-processing costs.
\newblock \emph{Proceedings of the Royal Society of London A: Mathematical,
  Physical and Engineering Sciences}, 469\penalty0 (2153), 2013.

\bibitem[Parisotto et~al.(2016)Parisotto, Ba, and Salakhutdinov]{ParisottoBS16}
Emilio Parisotto, Lei~Jimmy Ba, and Ruslan Salakhutdinov.
\newblock Actor-mimic: Deep multitask and transfer reinforcement learning.
\newblock \emph{International Conference on Learning Representations (ICLR)},
  2016.

\bibitem[Peters et~al.(2010)Peters, M{\"u}lling, and Alt{\"u}n]{Peters10}
Jan Peters, Katharina M{\"u}lling, and Yasemin Alt{\"u}n.
\newblock Relative entropy policy search.
\newblock In \emph{Proceedings of the Twenty-Fourth AAAI Conference on
  Artificial Intelligence (AAAI)}, 2010.

\bibitem[Rawlik et~al.(2012)Rawlik, Toussaint, and Vijayakumar]{rawlik12}
Konrad Rawlik, Marc Toussaint, and Sethu Vijayakumar.
\newblock On stochastic optimal control and reinforcement learning by
  approximate inference.
\newblock In \emph{(R:SS 2012)}, 2012.
\newblock \emph{Runner Up Best Paper Award}.

\bibitem[Rezende et~al.(2014)Rezende, Mohamed, and Wierstra]{rezende14}
Danilo~Jimenez Rezende, Shakir Mohamed, and Daan Wierstra.
\newblock Stochastic backpropagation and approximate inference in deep
  generative models.
\newblock In \emph{Proceedings of the 31st International Conference on Machine
  Learning (ICML)}, 2014.

\bibitem[Riedmiller et~al.(2018{\natexlab{a}})Riedmiller, Hafner, Lampe,
  Neunert, Degrave, Van~de Wiele, Mnih, Heess, and
  Springenberg]{riedmiller@neurocog}
Martin Riedmiller, Roland Hafner, Thomas Lampe, Michael Neunert, Jonas Degrave,
  Tom Van~de Wiele, Volodymyr Mnih, Nicolas Heess, and Jost~Tobias
  Springenberg.
\newblock Learning by playing – solving sparse reward tasks from scratch.
\newblock \emph{arXiv:1802.10567}, 2018{\natexlab{a}}.
\newblock URL \url{https://arxiv.org/abs/1802.10567}.

\bibitem[Riedmiller et~al.(2018{\natexlab{b}})Riedmiller, Hafner, Lampe,
  Neunert, Degrave, de~Wiele, Mnih, Heess, and
  Springenberg]{riedmiller2018learning}
Martin~A. Riedmiller, Roland Hafner, Thomas Lampe, Michael Neunert, Jonas
  Degrave, Tom~Van de~Wiele, Volodymyr Mnih, Nicolas Heess, and Jost~Tobias
  Springenberg.
\newblock Learning by playing - solving sparse reward tasks from scratch.
\newblock \emph{CoRR}, abs/1802.10567, 2018{\natexlab{b}}.
\newblock URL \url{http://arxiv.org/abs/1802.10567}.

\bibitem[Rubin et~al.(2012)Rubin, Shamir, and Tishby]{rubin2012trading}
Jonathan Rubin, Ohad Shamir, and Naftali Tishby.
\newblock \emph{Trading Value and Information in MDPs}, pp.\  57--74.
\newblock Springer Berlin Heidelberg, Berlin, Heidelberg, 2012.

\bibitem[Rusu et~al.(2016)Rusu, Colmenarejo, Gulcehre, Desjardins, Kirkpatrick,
  Pascanu, Mnih, Kavukcuoglu, and Hadsell]{rusu2016policy_distillation}
Andrei~A. Rusu, Sergio~Gomez Colmenarejo, Caglar Gulcehre, Guillaume
  Desjardins, James Kirkpatrick, Razvan Pascanu, Volodymyr Mnih, Koray
  Kavukcuoglu, and Raia Hadsell.
\newblock Policy distillation.
\newblock In \emph{International Conference on Learning Representations
  (ICLR)}, 2016.
\newblock URL \url{https://arxiv.org/abs/1511.06295}.

\bibitem[Schmitt et~al.(2018)Schmitt, Hudson, Zidek, Osindero, Doersch,
  Czarnecki, Leibo, Kuttler, Zisserman, Simonyan, and
  Eslami]{schmitt2018kickstarting}
Simon Schmitt, Jonathan~J. Hudson, Augustin Zidek, Simon Osindero, Carl
  Doersch, Wojciech~M. Czarnecki, Joel~Z. Leibo, Heinrich Kuttler, Andrew
  Zisserman, Karen Simonyan, and S.~M.~Ali Eslami.
\newblock Kickstarting deep reinforcement learning.
\newblock \emph{arXiv:1803.03835}, 2018.
\newblock URL \url{https://arxiv.org/abs/1803.03835}.

\bibitem[Schulman et~al.(2015)Schulman, Levine, Abbeel, Jordan, and
  Moritz]{schulman2015trust}
John Schulman, Sergey Levine, Pieter Abbeel, Michael Jordan, and Philipp
  Moritz.
\newblock Trust region policy optimization.
\newblock In \emph{Proceedings of the 32nd International Conference on Machine
  Learning (ICML-15)}, pp.\  1889--1897, 2015.

\bibitem[Schulman et~al.(2017{\natexlab{a}})Schulman, Abbeel, and
  Chen]{schulman2017equivalence}
John Schulman, Pieter Abbeel, and Xi~Chen.
\newblock Equivalence between policy gradients and soft q-learning.
\newblock \emph{arXiv preprint arXiv:1704.06440}, 2017{\natexlab{a}}.

\bibitem[Schulman et~al.(2017{\natexlab{b}})Schulman, Wolski, Dhariwal,
  Radford, and Klimov]{schulman2017ppo}
John Schulman, Filip Wolski, Prafulla Dhariwal, Alec Radford, and Oleg Klimov.
\newblock Proximal policy optimization algorithms.
\newblock \emph{CoRR}, abs/1707.06347, 2017{\natexlab{b}}.
\newblock URL \url{http://arxiv.org/abs/1707.06347}.

\bibitem[Simon(1956)]{simon1956rational}
{H.A.} Simon.
\newblock Rational choice and the structure of the environment.
\newblock \emph{Psychological Review}, 63:\penalty0 129--138, 1956.

\bibitem[Still \& Precup(2012)Still and Precup]{still2012information}
Susanne Still and Doina Precup.
\newblock An information-theoretic approach to curiosity-driven reinforcement
  learning.
\newblock \emph{Theory in Biosciences}, 131\penalty0 (3):\penalty0 139--148,
  2012.
\newblock \doi{10.1007/s12064-011-0142-z}.
\newblock URL \url{https://doi.org/10.1007/s12064-011-0142-z}.

\bibitem[Teh et~al.(2017)Teh, Bapst, Czarnecki, Quan, Kirkpatrick, Hadsell,
  Heess, and Pascanu]{teh2017distral}
Yee~Whye Teh, Victor Bapst, Wojciech~Marian Czarnecki, John Quan, James
  Kirkpatrick, Raia Hadsell, Nicolas Heess, and Razvan Pascanu.
\newblock Distral: Robust multitask reinforcement learning.
\newblock \emph{arXiv preprint arXiv:1707.04175}, 2017.

\bibitem[Tiomkin \& Tishby(2017)Tiomkin and Tishby]{tiomkin2017unified}
Stas Tiomkin and Naftali Tishby.
\newblock A unified bellman equation for causal information and value in markov
  decision processes.
\newblock \emph{CoRR}, abs/1703.01585, 2017.

\bibitem[Tishby \& Polani(2011)Tishby and Polani]{tishby2011information}
Naftali Tishby and Daniel Polani.
\newblock The information theory of decision and action.
\newblock In \emph{Percept. Action Cycle Springer Ser. in Cognitive Neural
  Syst.}, volume~19, pp.\  601--636, 01 2011.
\newblock ISBN 978-1-4419-1451-4.

\bibitem[Todorov(2007)]{todorov2007linearly}
Emanuel Todorov.
\newblock Linearly-solvable markov decision problems.
\newblock In B.~Sch\"{o}lkopf, J.~C. Platt, and T.~Hoffman (eds.),
  \emph{Advances in Neural Information Processing Systems 19}, pp.\
  1369--1376. MIT Press, 2007.

\bibitem[Toussaint(2009)]{toussaint2009robot}
Marc Toussaint.
\newblock Robot trajectory optimization using approximate inference.
\newblock In \emph{Proceedings of the 26th Annual International Conference on
  Machine Learning}, ICML '09, pp.\  1049--1056, 2009.
\newblock ISBN 978-1-60558-516-1.

\bibitem[Williams \& Peng(1991)Williams and Peng]{williams1991function}
R.~J. Williams and J.~Peng.
\newblock Function optimization using connectionist reinforcement learning
  algorithms.
\newblock \emph{Connection Science}, 3\penalty0 (3):\penalty0 241--268, 1991.

\bibitem[Zhang et~al.(2018)Zhang, Xiang, Hospedales, and Lu]{DML18}
Ying Zhang, Tao Xiang, Timothy Hospedales, and Huchuan Lu.
\newblock Deep mutual learning.
\newblock In \emph{Computer Vision and Pattern Recognition 2018}, 2018.

\bibitem[Ziebart(2010)]{ziebart2010modeling}
Brian~D. Ziebart.
\newblock \emph{Modeling Purposeful Adaptive Behavior with the Principle of
  Maximum Causal Entropy}.
\newblock PhD thesis, Machine Learning Department, Carnegie Mellon University,
  Dec 2010.

\end{thebibliography}
\bibliographystyle{iclr2019_conference}

\newpage
\appendix

\section{KL-regularized RL and information bottleneck}
\label{sec:appendix:IB}

In this appendix we derive the connection between KL-regularized RL and information bottleneck in detail. For simplicity we assume that $\xD_t$ is empty, consider dependence only on current state $s_t$ and do not use subscript by $t$ in detailed derivations for notational convenience. 
We also apologize for some notational inconsistencies, and will fix them in a later draft.

\subsection{Minimizing information flow from $S_t$ to $A_t$ for unstructured policies}

The simple formulation of the information bottleneck corresponds to maximizing reward while minimizing the per-timestep information between actions and state (or a subset of state, like the goal):

\begin{align}
\mathcal{L} &= \EE_{\qq}[\sum_t ( r(s_t, a_t) - \MutI[A_t;S_t]) ]
\end{align}

Upper-bounding the mutual information term:
\begin{align}
\MutI[A;S] &= \int {\qq}(s) {\qq}(a|s) \log \frac{{\qq}(s){\qq}(a|s)}{{\qq}(s) {\qq}(a)} \\
&= \int {\qq}(s) {\qq}(a|s) \log \frac{{\qq}(a|s)}{{\qq}(a)}\\
&\le \int {\qq}(s) {\qq}(a|s) \log \frac{{\qq}(a|s)}{\pp(a)}\\
&=\EE_{\qq} [ \KL [ {\qq}(A|s) \| \pp(A) | s] ] ,
\end{align}
since
\begin{align}
&0 \le{\qq} \KL[{\qq}(a) \| \pp(a)] = \EE_{\qq} [ \log \frac{{\qq}(a)}{\pp(a)} ] = \EE_{\qq} [ \log {\qq}(a)] - \EE_{\qq}[ \log \pp(a) ]\\
\iff &\EE_{\qq}[ \log {\qq}(a)] \ge \EE_{\qq}[ \log \pp(a) ].
\end{align}

Thus
\begin{align}
\mathcal{L} &= \EE_{\qq}[\sum_t ( r(s_t, a_t) - \MutI[A_t;S_t]) ] \\
& \ge \EE_{\qq}[\sum_t ( r(s_t, a_t) - \KL[ {\qq}_t \| \pp_t | s_t],
\end{align}
i.e.\ the problem turns into one of KL-regularized RL.

\subsection{Minimizing information flow from $S_t$ to $A_t$ for policies with latent variables}
\label{sec:appendix:IB:2}

For policies with latent variables such as ${\qq}(a|s) = \int {\qq}(a|z) {\qq}(z|s) dz$ we obtain:
\begin{align}
\MutI[A;S] & = \int {\qq}(a,s) \log {\qq}(a|s) da ds - \int {\qq}(a) \log {\qq}(a) da \\
&\le{\qq} \int {\qq}(a,s) \log {\qq}(a|s) da ds - \int {\qq}(a) \log \bar\pp(a) da
\end{align}
as before.

We choose $\pp(a) = \int {\qq}(a|z) \pp(z) dz$, then:
\begin{align}
\int {\qq}(a) \log \bar\pp(a) da &= \int {\qq}(a) \log \int {\qq}(a|z) \pp(z) dz da\\
&= \int {\qq}(a,s) \log \int {\qq}(a|z) \pp(z) dz da ds\\
&= \int {\qq}(a,s) \log \int {\qq}(a|z) \frac{{\qq}(z|s,a)}{{\qq}(z|s,a)} \pp(z) dz da ds\\
&\geq \int {\qq}(a,s,z) \log \frac{{\qq}(a|z)\pp(z)}{{\qq}(z|s,a)} dz da ds\\
&= \int {\qq}(a,s,z) \log \frac{{\qq}(a|z)\pp(z){\qq}(a|s)}{{\qq}(a|z){\qq}(z|s)} dz da ds\\
&= \int {\qq}(a,s) \log {\qq}(a|s)da ds + \int {\qq}(z,s)\log \frac{\pp(z)}{{\qq}(z|s)} dz ds,
\end{align}
and thus
\begin{align}
\MutI[A;S] & \leq \int {\qq}(a,s) \log q(a|s) da ds - \int {\qq}(a) \log \bar\pp(a) da \\
&\leq \int {\qq}(a,s) \log {\qq}(a|s) da ds -
 \int {\qq}(a,s) \log {\qq}(a|s)da ds - \int {\qq}(z,s)\log \frac{\pp(z)}{{\qq}(z|s)} dz \\
&= \int {\qq}(z,s)\log \frac{{\qq}(z|s)}{\pp(z)}dz = \EE_{\qq}[ \KL[{\qq}(Z|s) \| \pp(Z) | s ] ].
\end{align}
Therefore:
\begin{align}
\mathcal{L} &= \EE_{\qq}[\sum_t ( r(s_t, a_t) - \MutI[A_t;S_t]) ] \\
& \geq \EE_{\qq}[\sum_t ( r(s_t, a_t) - \KL[ {\qq}(Z_t | s) \| \pp(Z_t) | s_t],
\end{align}

Thus, the KL regularized objective discussed above can be seen as implementing an information bottleneck. Different forms of the default policy correspond to restricting the information flow between different components of the interaction history (past states or observations), and to different approximations to the resulting mutual information penalties.

This perspective suggests two different interpretations of the KL regularized objective discussed above: We can see the role of the default policy implementing a way of restricting information flow between (past) states and (future) actions. An alternative view, more consistent with the analogy between RL and probabilistic modeling invoked above is that of learning a ``default'' behavior that is independent of some aspect of the state. (Although the information theoretic view has recently gained more hold in the probabilistic modeling literature, too \citep[e.g.][]{alemi2016deep,alemi2017fixing}).

\section{Distributed learning setup}
\label{sec:appendix:distributed}

\begin{figure}[!htp]
    \begin{center}
        \includegraphics[scale=0.15]{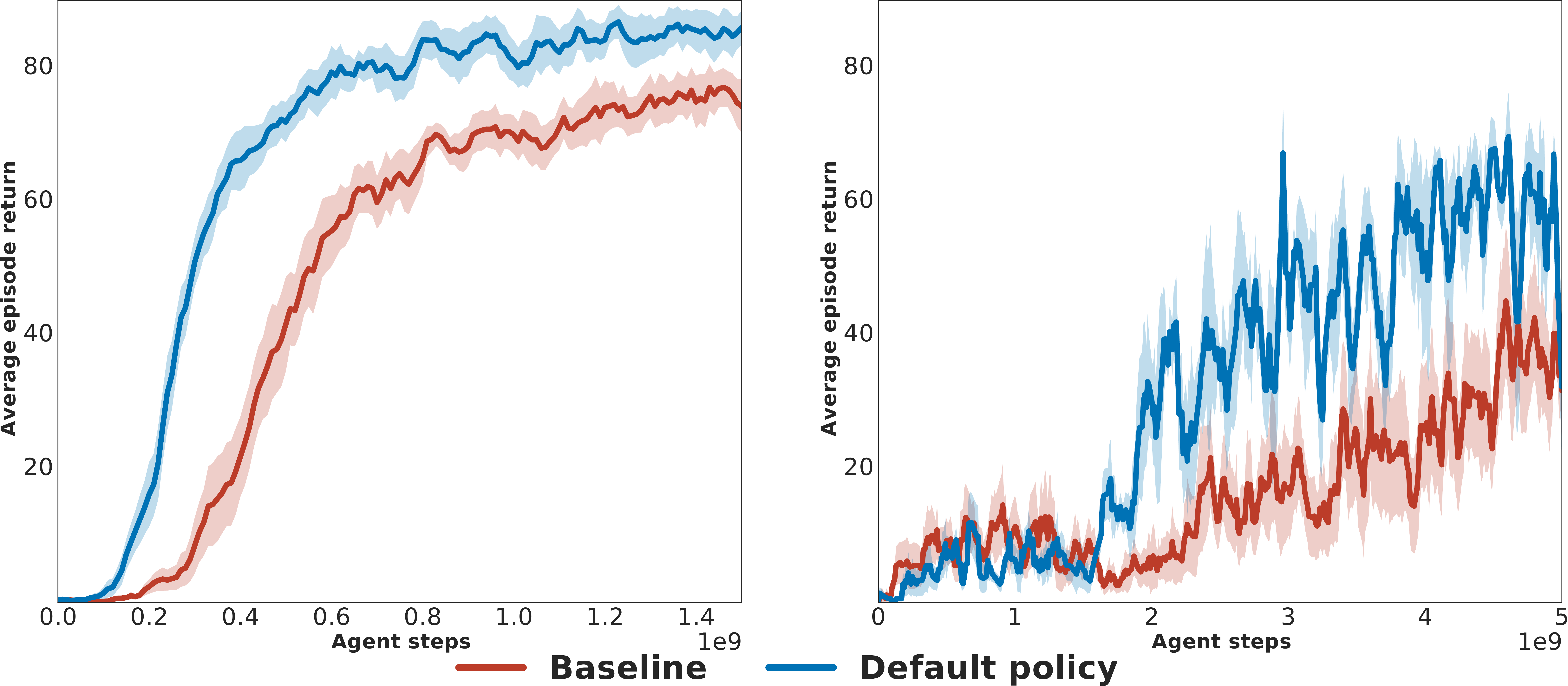}
        \caption{\textbf{Single versus multiple actors} comparison on go to moving target task. \textbf{Left}: 32 actors. \textbf{Right}: 1 actor.}
        \label{fig:appendix:distributed}
    \end{center}
\end{figure}

We use a distributed off-policy setup similar to~\citet{riedmiller@neurocog}. There is one learner and multiple actors. These are essentially the instantiations of the main agent used for different purposes. Each actor is the main agent version which receives the copy of parameters from the learner and unrolls the trajectories in the environment, saving it to the replay buffer of fixed size $1e6$. The learner is the agent version which samples a batch of short trajectories windows (window size is defined by \textit{unroll length}) from the replay buffer, calculates the gradients and updates the parameters. The updated parameters are then communicated to the actors. Such a setup speeds-up learning significantly and makes the final performance of the policy better. We compare the performance of on \textit{go to moving target task} with 1 and 32 actors. From figure~\ref{fig:appendix:distributed}, we see that the effect of the default policy does not disappear when the number of actor decreases to 1, but the learning becomes much slower, noisier and weaker.

\section{Continuous control: Walkers and task details}
\label{sec:appendix:walkers_tasks}

\begin{figure}[!h]
    \begin{center}
        \subfigure[\textit{Jumping ball}]{\label{fig:walkers:bb8}\includegraphics[scale=0.2]{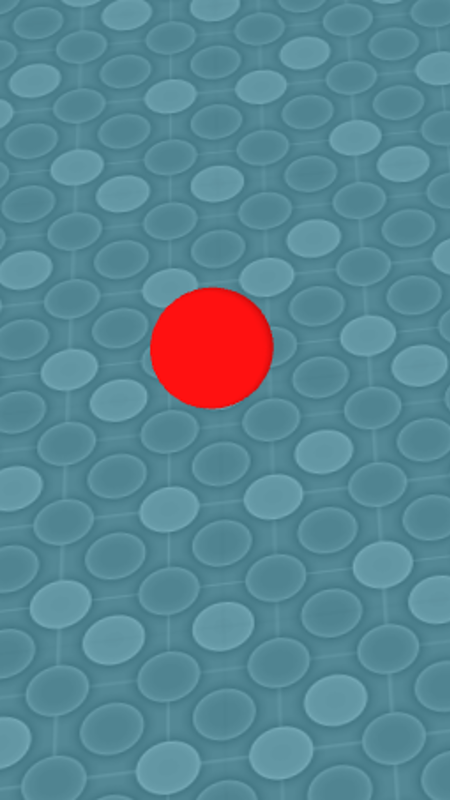}}
        \subfigure[\textit{Quadruped}]{\label{fig:walkers:ant}\includegraphics[scale=0.2]{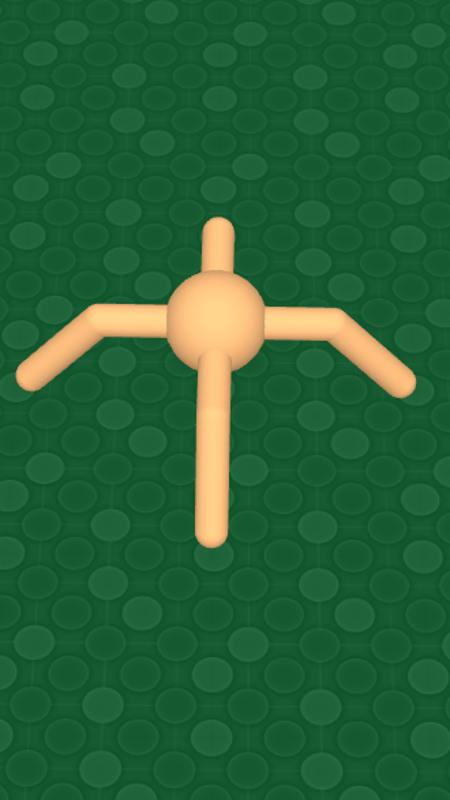}}
        \subfigure[\textit{Humanoid}]{\label{fig:walkers:humanoid}\includegraphics[scale=0.2]{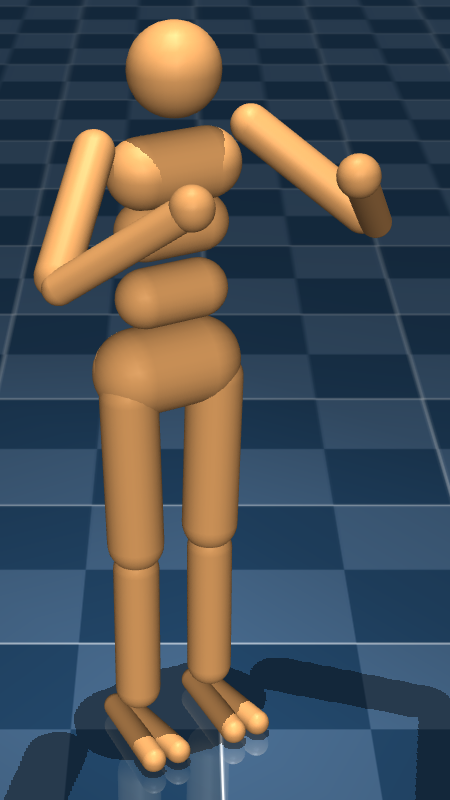}}
        \caption{\textbf{Walkers visualization}. }
        \label{fig:walkers}
    \end{center}
\end{figure}

Walkers visualization is provided in figure~\ref{fig:walkers}. Below we give a detaatiled description of each continuous control task we studied.

\textbf{Walking task}.\\
\textit{Type}. \textit{Dense-reward feature-based-task}.\\
\textit{Description}. Each half of the episode, a random direction among 4 (left, right, forward and backwards) is sampled.
Task information is specified via a one-hot encoding of the required direction. The walker is required to move in this direction with the target speed $v_{t}$ and receives the reward $r$.\\
\textit{Reward}. $r=\exp^{ -|v_{cur} - v_{t}|^{2}}$.\\
\textit{Technical details}. Target speed, $v_{t} = 3$. The episode length is 10 seconds. For the humanoid task we use the absolute head height termination criteria: $h < 0.95$.

\textbf{Walls}.\\
\textit{Type}. \textit{Dense-reward vision-task}.\\
\textit{Description}. Walker is required to run through a terrain and avoid the walls. The task-specific information is a vision input. It receives the reward $r$ defined as a difference between the current walker speed $v_{cur}$ and the target speed $v_{t}$ along the direction of the track.\\
\textit{Reward}. $r=\exp^{ -|v_{cur} - v_{t}|^{2}}$.\\
\textit{Technical details}. Target speed, $v_{t} = 3$. The episode length is 45 seconds. For the humanoid task we use the absolute head height termination criteria: $h < 0.9$.

\textbf{Go to one of K single targets}.\\
\textit{Type}. \textit{Sparse-reward feature-based-task}.\\
\textit{Description}. On an infinite floor, there is a finite area of size 8x8 with K randomly placed targets. The walker is also randomly placed in a finite area. The walker's initial position is also randomly placed on the finite area. The walker is required to one of the K targets, specified via command vector. Once it achieves the target, the episode terminates and the walker receives the reward $r$.\\
\textit{Reward}. $r = 60$.\\
\textit{Technical details}. The episode length is 20 seconds.

\textbf{Go to one moving target}.\\
\textit{Type}. \textit{Sparse-reward feature-based-task}.\\
\textit{Description}. Similar to the previous one, but there is only one target and once the walker achieves it, the target reappears in a new random place. The walker receives $r$ for 10 consecutive steps staying on the target before the target reappears in a new random position.\\
\textit{Reward}. $r=1$.\\
\textit{Technical details}. The episode length is 25 seconds.

\textbf{Move one box to one of the K targets}.\\
\textit{Type}. \textit{Sparse-reward feature-based-task}.\\
\textit{Description}. There is a finite floor of size 3x3 padded with walls with K randomly placed targets and one box. The walker is required to move this box to one of the specified targets. Once the box is placed on the target, the episode terminates and the walker receives the reward $r$.\\
\textit{Reward}. $r=60$. \\
\textit{Technical details}. The episode length is 30 seconds. Control timestep is 0.05 for \textit{quadruped} and 0.025 for \textit{jumping ball}.

\textbf{Move one box to one of the K targets and go to another}.\\
\textit{Type}. \textit{Sparse-reward feature-based-task}.\\
\textit{Description}. Similar to the previous one, but the walker is also required to go to another target (which is different from the one where it must place the box on). The walker receives the a $r_{task}$ for each task solved, and a $r_{end}$ if it solves both tasks. The other parameters are the same.\\
\textit{Reward}. $r_{task} = 10$, $r_{end}=50$.\\
\textit{Technical details}. Same as in the previous task.

\textbf{Foraging in the maze}.\\
\textit{Type}. \textit{Sparse-reward vision-task}.\\
\textit{Description}. There is a maze with 8 apples which walker must collect. For each apple, it receives reward $r$. The episode terminates once the walker collects all the apples or the time is elapsed. \\
\textit{Reward}. $r=1$.\\
\textit{Technical details}. The episode length is 90 seconds. Control timestep is 0.025 for \textit{jumping ball}, and 0.05 for \textit{quadruped}.

\section{Algorithms, baseline and hyperparameters}
\label{sec:appendix:algorithm}

Our agents run in off-policy regime sampling the trajectories from the replay buffer. In practice, it means that the trajectories are coming from the behavior (replay buffer) policy $\pi_{b}$, and thus, the correction must be applied (specified below). Below we provide architecture details, baselines, hyperparmaeters as well as algorithm details for discrete and continuous control cases. 

\subsection{Discrete case}
\label{sec:appendix:algorithm:discrete}

In discrete experiments, we use V-trace off-policy correction as in ~\cite{espeholt2018impala}. We reuse all the hyperparameters for DMLab from the mentionned paper. At the top of that, we add default policy network and optimize the corresponding $\alpha$ parameter using population-base training. The difference with the setup in ~\cite{espeholt2018impala} is that they use the human prior over actions (table D.2 in the mentionned paper), which results in 9-dimensional action space. In our work, we take the rough DMLab action space, consisting of all possible rotations, and moving forward/backward, and "fire" actions. It results in the action space of dimension 648. It make the learning much more challenging, as it has to explore in much larger space.

\subsection{Continuous case}
\label{sec:appendix:algorithm:continuous}

The agent network (see figure~\ref{fig:prior_posterior_architecture}) is divided into actor and critic networks without any parameter sharing. In the case of \textit{feature-based-task}, the task-specific information is encoded by one layer MLP with ELU activations. For the \textit{vision-task}, we use a 3-layer ResNet~\cite{he2015resnet}. The encoded task information is then concatenated with the proprioceptive information and passed to the agent network. The actor network encodes a Gaussian policy, $\mathcal{N}(\tilde{\mu}, \tilde{\sigma})$, by employing a two-layer MLP, with mean $\mu$ and log variance $\log \sigma$ as outputs and applying the following processing procedures:
\begin{equation*}
    \tilde{\mu} = tanh(\mu),
\end{equation*}
\begin{equation*}
    \tilde{\sigma} = 0.1 + (\sigma_{max} - 0.1) f (\log\sigma),
\end{equation*}
where $f$ is a sigmoid function:
\begin{equation*}
    f(x) = \frac{1}{1 + \exp^{-x}}.
\end{equation*}

The critic network is a two-layer MLP and a linear readout. The default policy network has the same structure as actor network, but receives a concatenation of the proprioceptive information with only a subset (potentially, empty) of a task-specific information. There is no parameter sharing between the agent and the default policy. ELU is used as activation everywhere. The exact actor, critic and default policy network architectures are described below. We tried to use LSTM for the default policy network instead of MLP, but did not see a difference. We use separate optimizers and learning rates $\beta_{\qq}, \beta_{Q}, \beta_{\pp}$ for the actor, critic and default policy networks correspondingly. For each network (which we call \textit{online}), we also define the \textit{target} network, similar to the target $Q$-networks ~\citep{mnih2015human}. The target networks are updated are updated in a slower rate than the online ones by copying their parameters.

We assume that the trajectories are coming from the replay buffer $\mathcal{B}$. To correct for being off-policy, we make use of the Retrace operator (see~\cite{MunosSHB16}). This operator is applied to the $Q$ function essentially introducing the importance weights. We will note $\mathcal{R}Q$ the action for this operator. Algorithm~\ref{alg:KStep_op_cc} is an off-policy version with retraced $Q$ function of the initial algorithm~\ref{alg:KStep}.

\begin{algorithm}
\caption{Off-policy corrected version of algorithm~\ref{alg:KStep} for continuous control}
\label{alg:KStep_op_cc}
\begin{algorithmic}
    \State online policy: $\qq_{O, \theta_{O}}$, initial parameters $\theta_{O}$
    \State target policy: $\qq_{T, \theta_{T}}$, initial parameters $\theta_{T}$
    \State online default policy:  $\pp_{O, \phi_{O}}$; initial parameters $\phi_{O}$
    \State target default policy:  $\pp_{T, \phi_{T}}$; initial parameters $\phi_{T}$
    \State online Q-function:  $Q_{O, \psi_{O}}$; initial parameters $\psi_{O}$
    \State target Q-function:  $Q_{T, \psi_{T}}$; initial parameters $\psi_{T}$
    \State target update period: $P$
    \State replay buffer: $\mathcal{B}$
    \State unroll length: $K$
    \For{j=1, \dots}
        \State Sample partial trajectory from replay buffer $\mathcal{B}$: $\tau_{t:t+K} = (s_t, a_t, r_t \dots r_{t+K}) $
        \State compute online KL: $\widehat{\KL}_{O, t'} = \KL[ \qq_{O}(\cdot | s_{t'}) \| \pp_{O}(\cdot | s_{t'}) ]$
        \State compute target KL: $\widehat{\KL}_{T, t'} = \KL[ \qq_{O}(\cdot | s_{t'}) \| \pp_{T}(\cdot | s_{t'}) ]$
        \State Estimate boostrap value: $\hat{V} = \EE_{\qq_{T}(\cdot|s_{t+K})}[ Q_{T}(s_{t+K}, a) ] - \alpha \widehat{\KL}_{T, t+K}$
        \State Estimate Q targets: $\hat{Q}_{t'} = r_{t'} +  \hat{V}$
        \State Apply Retrace operator: $\hat{Q}^{R}_{t'} = \mathcal{R}\hat{Q}_{t'}$
        \State Agent policy loss: $\hat{L}_\qq = \sum_{t'=t}^{t+K-1} \EE_{\qq_{O}(\cdot| s_{t'})} [ Q_{T}(s_{t'}, a) ] - \alpha \widehat{KL}_{T, t'}$
        \State Q-value loss: $\hat{L}_Q  = \sum_{t'=t}^{t+K-1} \| \hat{Q}^{R}_{t'} - Q_{O}(s_{t'}, a_{t'}) \|^2$
        \State Default policy loss: $\hat{L}_{\pp} = \sum_{t'=t}^{t+K-1} \widehat{\KL}_{O, t'}$
        \State $\theta_{O} \leftarrow \theta_{O} + \beta_{\qq} \nabla_\theta \hat{L}_{\qq}$
        \State $\phi_{O} \leftarrow \phi_{O} + \beta_{\pp} \nabla_{\phi} \hat{L}_{\pp}$
        \State $\psi_{O} \leftarrow \psi_{O} - \beta_{Q} \nabla_{\psi} \hat{L}_{Q}$
        \If{$j \mod P = 0$}
            \State $\theta_{T} \leftarrow \theta_{O}$
            \State $\phi_{T} \leftarrow \phi_{O}$
            \State $\psi_{T} \leftarrow \psi_{O}$
        \EndIf
    \EndFor
\end{algorithmic}
\end{algorithm}

We use the same update period for actor and critic networks, $P_{a}$ and a different period for the default network $P_{d}$. The baseline is the agent network (see figure~\ref{fig:prior_posterior_architecture}) without the default policy with an entropy bonus $\lambda$. All the hyperparameters of the baseline are tuned for each task. For each best baseline hyperparameters configuration, we tune the default policy parameters. When we use the default policy, we do not have the entropy bonus. Instead, we have a \textit{regularisation parameter} $\alpha$. The other parameteres which we consider are: \textit{batch size}, \textit{unroll length}. Below we provide the hyperparameters for each of the task. The following default hyperparameters are used unless some particular one is specified.

\textbf{Default hyperparameters}.\newline
\textit{Actor learning rate}, $\beta_{\qq} = 0.0005$. \newline
\textit{Critic learning rate}, $\beta_{Q} = 0.0005$. \newline
\textit{Default policy learning rate}, $\beta_{\pp} = 0.0005$. \newline
\textit{Agent target network update period}: $P_{a} = 100$. \newline
\textit{Default policy target network update period}: $P_{d} = 100$. \newline
\textit{Actor network}: MLP with sizes $(300, 200)$. \newline
\textit{Critic network}: MLP with sizes $(400, 300)$. \newline
\textit{Default policy network}: MLP with sizes $(300, 200)$. \newline
\textit{Command encoder network}: 1-layer MLP of size $50$. \newline
\textit{Image encoder}: ResNet with filter sizes $(16, 32, 32)$. \newline
\textit{Gaussian policy maximum noise}: $\sigma_{max}=1.0$. \newline
\textit{Batch size}: 512. \newline
\textit{Unroll length}: 10. \newline
\textit{Entropy bonus}: $\lambda = 0.0001$. \newline
\textit{Regularization constant}: $\alpha=0.01$. \newline
\textit{Number of actors}: 128.

\textbf{Walking \textit{quadruped}}\newline
\textit{Actor network}: MLP with sizes $(400, 300, 200)$. \newline
\textit{Critic network}: MLP with sizes $(400, 400, 300)$. \newline
\textit{default policy network}: MLP with sizes $(400, 300, 200)$. \newline
\textit{Regularization constant}: $\alpha=0.0001$. \newline
\textit{Number of actors}: 256.

\textbf{Walking \textit{humanoid}}\newline
\textit{Entropy bonus}: $\lambda = 0.005$. \newline
\textit{Regularization constant}: $\alpha=0.0001$.\newline
The rest is similar to \textbf{Walking \textit{quadruped}}.

\textbf{Walls \textit{quadruped}}\newline
\textit{Actor and critic learning rate}: $\beta_{\qq}, \beta_{Q} =5e-5$. \newline
\textit{Batch size}: 48. \newline
\textit{Regularization constant}: $\alpha=0.001$.\newline
\textit{Number of actors}: 64.

\textbf{Walls \textit{humanoid}}\newline
\textit{Actor and critic learning rate}: $\beta_{\qq}, \beta_{Q} =0.0001$. \newline
\textit{Batch size}: 48. \newline
\textit{Regularization constant}: $\alpha=0.001$.\newline
\textit{Number of actors}: 64.

\textbf{Go to moving target \textit{quadruped}}\newline
\textit{Regularization constant}: $\alpha=0.006$.\newline
\textit{Number of actors}: 32.

\textbf{Go to moving target \textit{humanoid}}\newline
\textit{Regularization constant}: $\alpha=0.1$.

\textbf{Go to K targets \textit{quadruped}}\newline
\textit{Actor and critic learning rate}: $\beta_{\qq}, \beta_{Q} =0.0001$. \newline
\textit{default policy target network update period}: $P_{d} = 50$. \newline
\textit{Regularization constant}: $\alpha=0.006$.

\textbf{Move 1 box to 1 target \textit{jumping ball}}\newline
Default

\textbf{Move 1 box to 1 target \textit{quadruped}}\newline
\textit{Actor and critic learning rate}: $\beta_{\qq}, \beta_{Q} =0.0001$.

\textbf{Move 1 box to one of 2 targets \textit{quadruped}}\newline
\textit{Actor and critic learning rate}: $\beta_{\qq}, \beta_{Q} =0.0001$. \newline
\textit{default policy target network update period}: $P_{d} = 50$.

\textbf{Move 1 box to one of 2 targets with go to another one \textit{quadruped}}\newline
Same as previous task.

\textbf{Move 1 box to one of 3 targets \textit{quadruped}}\newline
Same as previous task.

\textbf{Foraging \textit{jumping ball}}\newline
\textit{Actor and critic learning rate}: $\beta_{\qq}, \beta_{Q} =0.0001$. \newline
\textit{Actor network}: LSTM with one hidden unit of size $128$. \newline
\textit{Critic network}: LSTM with one hidden unit of size $128$. \newline
\textit{Batch size}: 48.\newline
\textit{Number of actors}: 64.

\textbf{Foraging \textit{quadruped}}\newline
\textit{Unroll length}: 20 \newline
\textit{Regularization constant}: $\alpha=0.006$.

For the \textbf{Foraging} \textit{quadruped} task, the initial agent did not learn, so we used a slightly different version of the agent. In algorithm~\ref{alg:KStep_op_cc}, we essentially learn a $Q$ function and update the policy by sampling actions from it and backpropagating through $Q$. In this algorithm, we learn a value function $V$ using V-trace~\citep{espeholt2018impala} and the policy is updated using an off-policy corrected policy gradient with empirical returns.

\section{Continuous Control: Additional results}
\label{sec:appendix:additional_results}

\subsection{Dense reward tasks}
\label{sec:appendix:additional_results:dense_reward}

\begin{figure}[!htp]
    \begin{center}
        \includegraphics[scale=0.09]{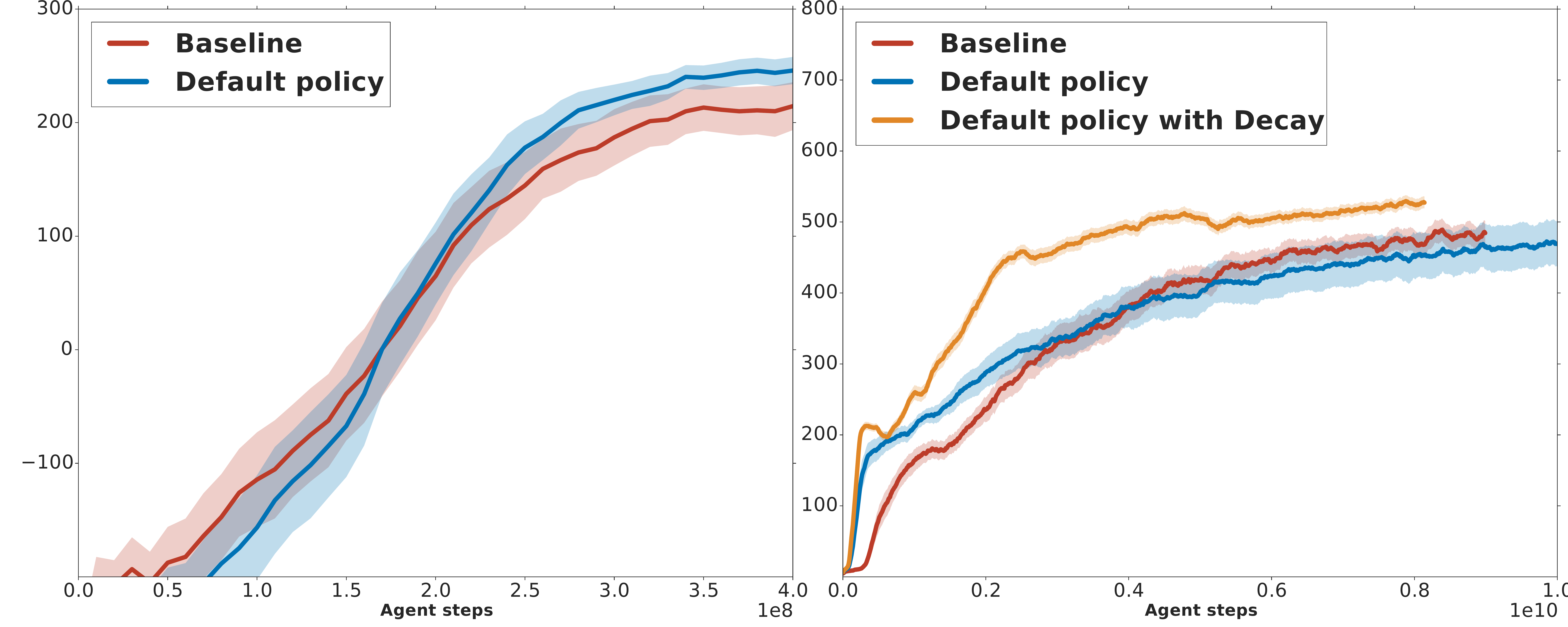}
        \caption{\textbf{Shaping reward} on \textit{go to moving target} task. \textbf{Left}: \textit{Quadruped} body. \textbf{Right}: \textit{Humanoid} body.}
        \label{fig:shaping_reward}
    \end{center}
\end{figure}

\begin{figure}[!htp]
    \begin{center}
        \includegraphics[scale=0.085]{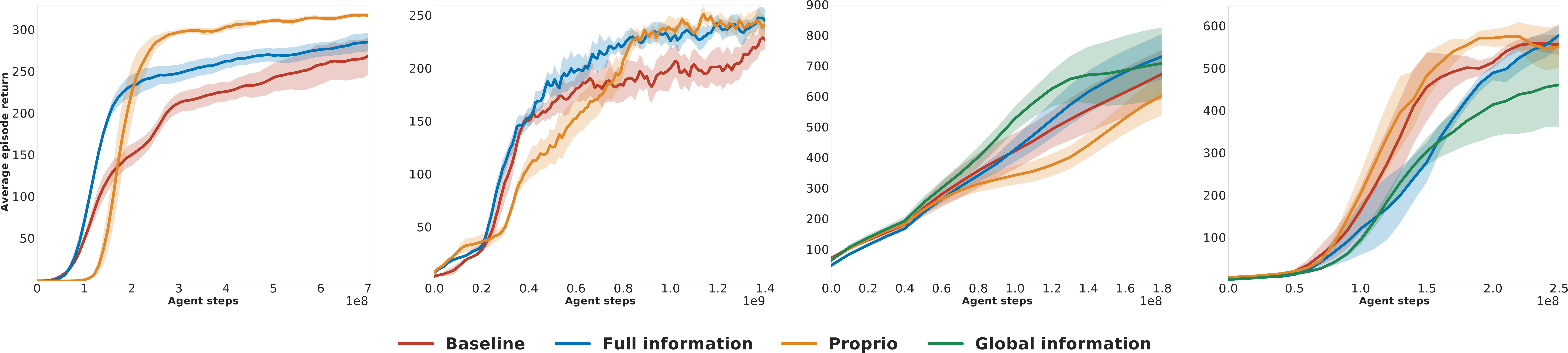}
        \caption{
            \textbf{Results for the dense-reward tasks}. Starting from left. \textbf{First}: walking \textit{quadruped} task. \textbf{Second}: walking \textit{humanoid} task. \textbf{Third}: walls \textit{quadruped} task. \textbf{Forth}: walls \textit{humanoid} task.
            The legend indicates the information available to the default policy (except for \emph{baseline} which does not use a default policy). The global information represents global position and orientation of the walker.
        }
        \label{fig:appendix:results:dense_reward}
    \end{center}
\end{figure}

In order to understand in more detail the regularization provided by the default policy we study its effect in dense reward tasks. In Figure~\ref{fig:shaping_reward} we show results for the \textit{go to moving target} task when the reward provided is proportional to the inverse distance between the walker and the target. For the \textit{quadruped} body, the gain provided by the default policy disappears (Fig.~\ref{fig:shaping_reward} left). For the humanoid body regularization against the default policy appears to provide an initial gain but subsequently slows down learning. This may be due to the fact that the humanoid typically first learns to stand (corresponding to a reward of roughly 200) before beginning to navigate to the target. The regularization provided by the default policy may slow down the second stage of learning. This is supported by the fact that decaying the influence of the default policy as learning progresses leads to gains compared to both fixed regularization and baseline. 
Figure~\ref{fig:appendix:results:dense_reward} shows results for additional tasks with dense rewards. Overall, benefits of the default policy are much less clear cut or absent compared to the \textit{sparse reward} case. One possible explanation is that the dense reward provides a strong learning signal everywhere in state space. Thus the policy learns quickly while the default policy lags behind and the benefit of generalization across the state space is limited. Also, a dense and hence more ``prescriptive'' reward may be less compatible with a default behavior.

\subsection{Sparse reward jumping ball}
\label{sec:appendix:additional_results:sparse_reward:bb8}

The results for the sparse reward tasks with \textit{jumping ball} are given in figure~\ref{fig:appendix:results:dense_reward}. We observe little difference when using a default policy compared to the baseline. We hypothesize that for the simple action space (3 action dimensions) considered in these experiments the regularization effect of the default policy on the state-conditional action distribution is limited. .

\begin{figure}[!htp]
    \begin{center}
        \includegraphics[scale=0.11]{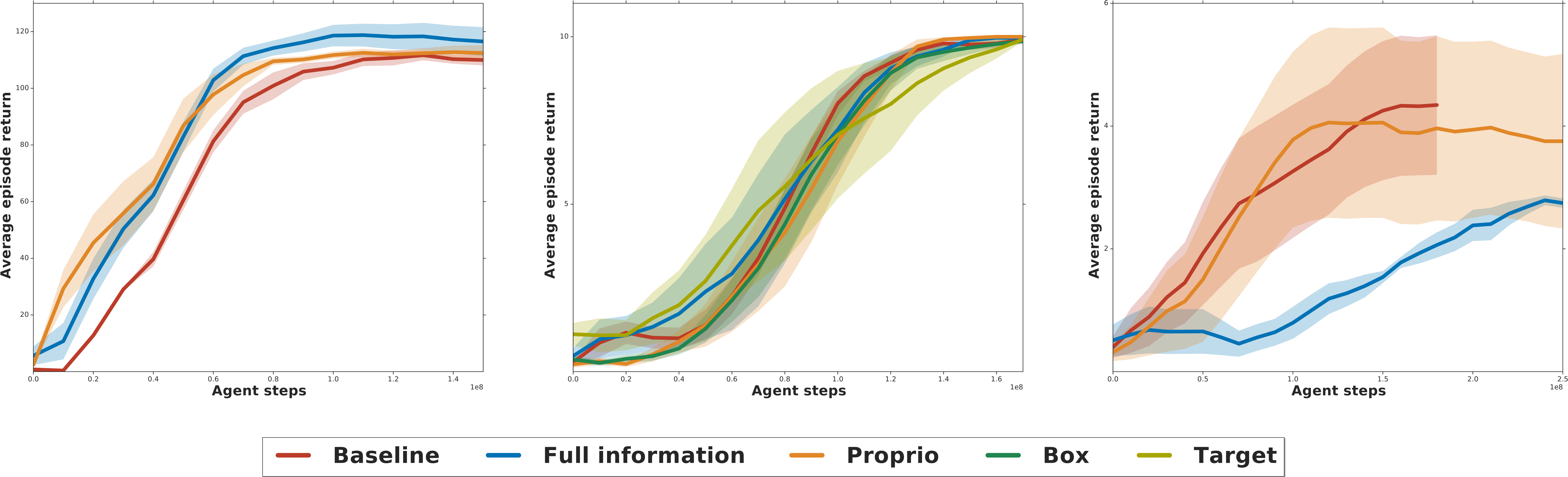}
        \caption{
            \textbf{Results for sparse-reward tasks} with \textit{jumping ball} walker.\textbf{Left}: go to moving target. \textbf{Center}: moving one box to one target. \textbf{Right}: foraging in the maze. The legends denote additional to the proprioception, information passed to the default policy (except baseline, where we do not use default policy).}
        \label{fig:appendix:results:sparse_reward:bb8}
    \end{center}
\end{figure}

\subsection{Sparse reward with quadruped additional results}
\label{sec:appendix:additional_results:sparse_reward:ant}

\begin{figure}[!htp]
    \begin{center}
        \includegraphics[scale=0.11]{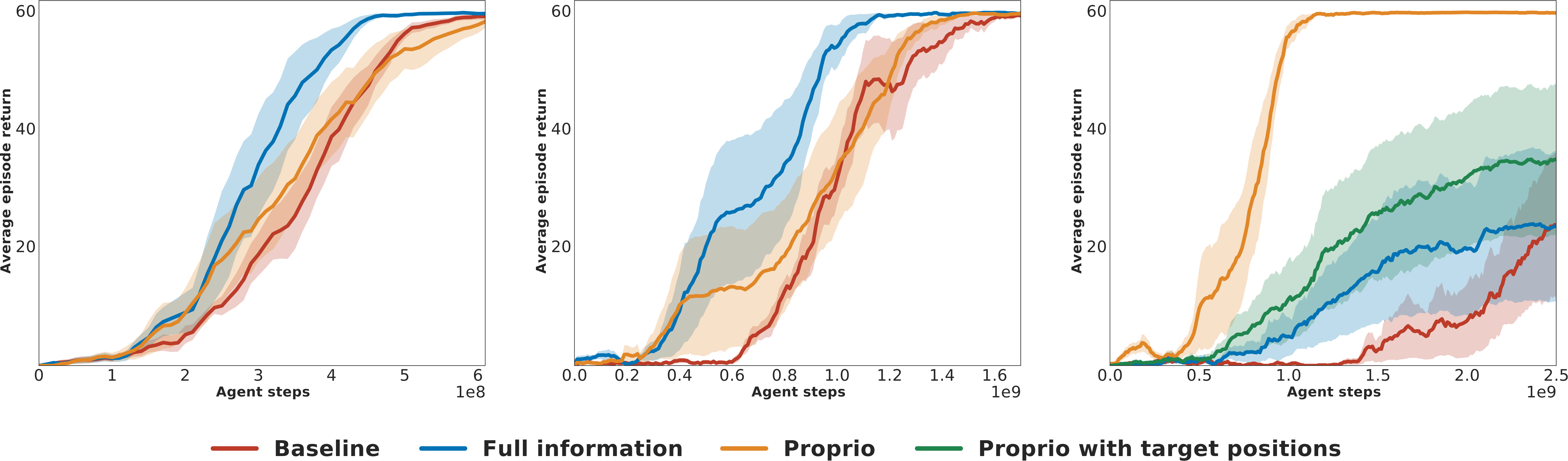}
        \caption{
            \textbf{Results for go to one of K targets tasks} with \textit{quadruped}. \textbf{Left}: go to 1 target. \textbf{Center}: go to one of 2 targets. \textbf{Right}: go to one of 3 targets. The legends denote additional to the proprioception, information passed to the default policy (except baseline, where we do not use default policy).
        }
        \label{fig:appendix:results:dense_reward:gtt_k}
    \end{center}
\end{figure}

In this section, we provide more results for the sparse reward tasks. In figure~\ref{fig:appendix:results:dense_reward:gtt_k} the results for going to one of K targets task with \textit{quadruped} are presented. The proprioceptive default policy gives significant gains comparing to others. What interesting is that when the number of targets $K$ increases, the baseline performance drops dramatically, whereas the proprioceptive default policy solve the task reliably. Our hypothesis is that the default policy learns quickly the default walking behavior which becomes very helpful for the agent to explore the floor and search for the target.

\begin{figure}[!htp]
    \begin{center}
        \includegraphics[scale=0.085]{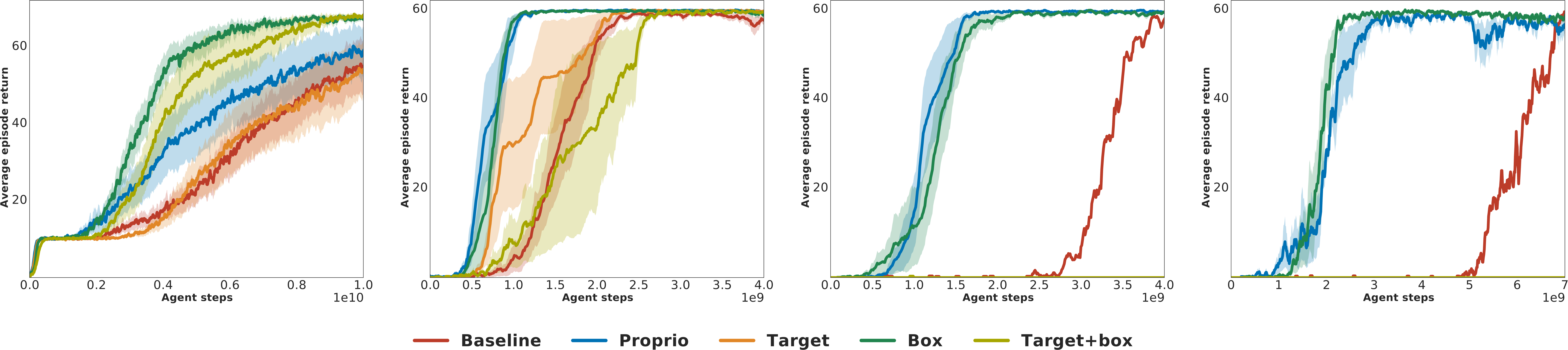}
        
        \caption{
            \textbf{Results for box pushing tasks} with \textit{quadruped}. Starting from left, \textbf{first}: move one box to one of 2 targets with go to another. \textbf{Second}: move one box to 1 target. \textbf{Third}: move one box to one of 2 targets. \textbf{Forth}: move one box to one of 3 targets. The legends denote additional to the proprioception, information passed to the default policy (except baseline, where we do not use default policy).
        }
        \label{fig:appendix:results:dense_reward:mb_k}
    \end{center}
\end{figure}

We also provide the results for move box to one of K targets task, where $K=1,2,3$, and move box to one of two targets task with go to another. The results are given in figure~\ref{fig:appendix:results:dense_reward:mb_k}. Similar effect occurs here.

\subsection{Additional transfer results}
\label{sec:appendix:additional_results:transfer}

\begin{figure}[!htp]
    \begin{center}
        \includegraphics[scale=0.1]{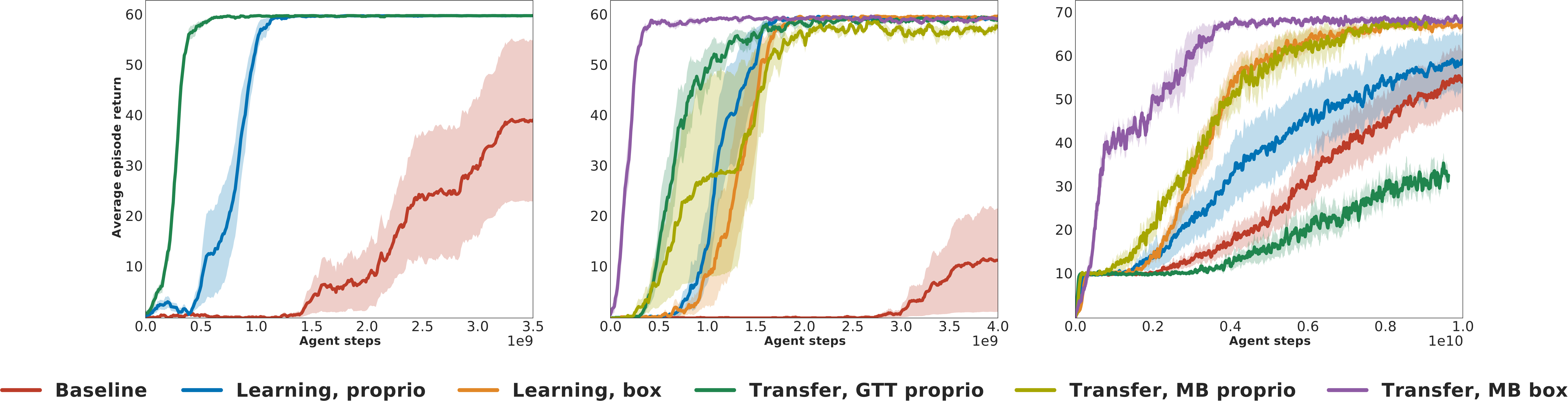}
        \caption{
            \textbf{Performance of the transfer} with \textit{quadruped} walker. \textbf{Left}: Go to one of 3 targets. \textbf{Center}: move one box to one of two targets. \textbf{Right}: move one box to one of two targets and go to another one. The legend whether the default policy is learned or is transferred. Furthermore, it specifies the task from which the default policy is transferred as well as additional information other than the proprioceptive information that the default policy is conditioned on, if any.
        }
        \label{fig:appendix:transfer}
    \end{center}
\end{figure}

In this section, we provide additional transfer experiment results for the range of the tasks. They are given in figure~\ref{fig:appendix:transfer}. In the first two cases we see that proprioceptive default policy from \textit{the go to target task} gives a significant boost to the performance comparing to the learning from scratch. We also observe, that for the \textit{box pushing tasks}, the default policy with the box position significantly speeds up learning comparing to other cases. We believe it happens because this default policy learns the best default behavior for these tasks possible: going to the box and push it. For the most complicated task, \textit{move one box to one of two targets and go to another one},~\ref{fig:appendix:transfer}, right, the box default policy makes a big difference: it makes the policy avoid being stuck in go to target behavior (line with reward of 10).

Additional results for the transfer experiments are given in figure~\ref{fig:appendix:transfer}. We observe the same effect happening: whereas the baseline performance drops significantly, the agent with default policy stays 

\subsection{Ablation Walls quadruped}
\label{sec:appendix:additional_results:ablations:walls:quadruped}

\begin{figure}[!htp]
    \begin{center}
        \includegraphics[scale=0.11]{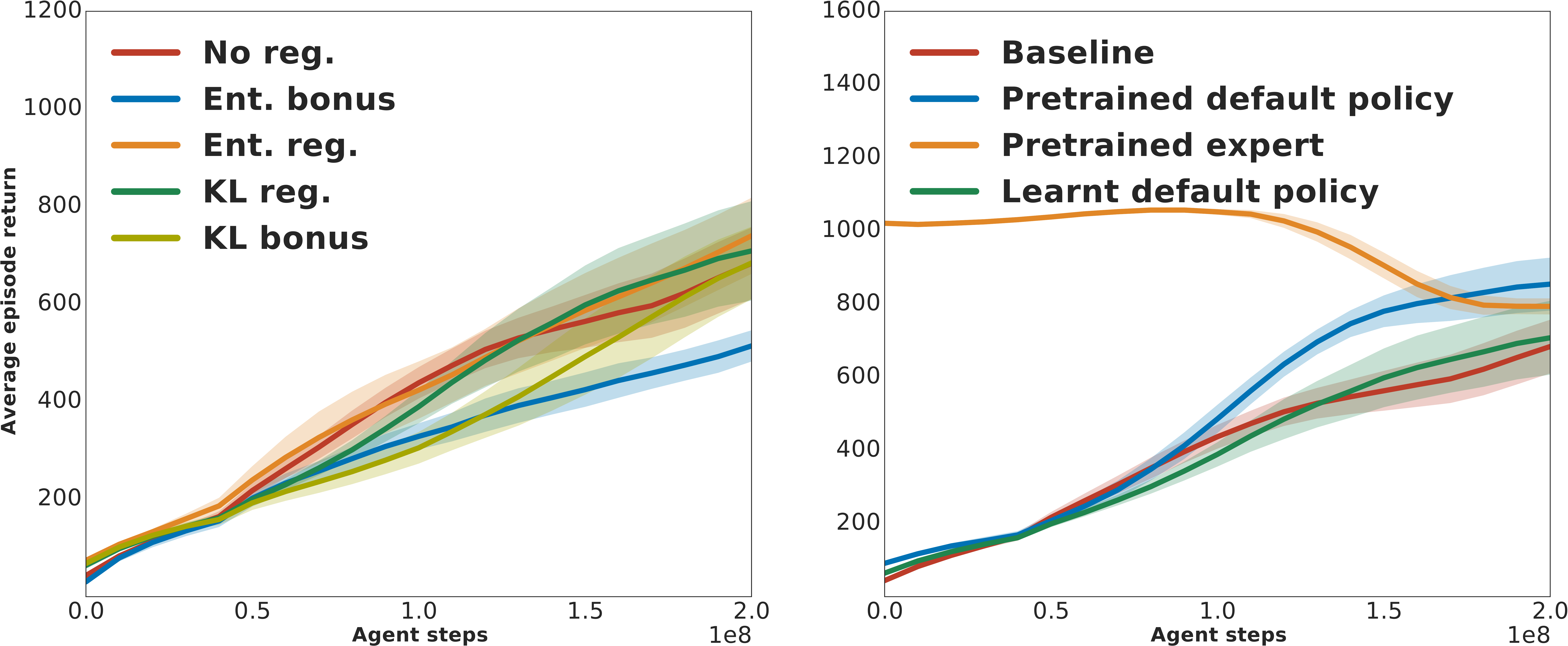}

        \caption{\textbf{Ablations} for walls task with \textit{quadruped}.  \textbf{Left}: Comparing various regularization schemes. \textbf{Right}: Optimistic baselines comparing pretrained default policies.
        }
        \label{fig:appendix:ablations:walls:ant}
    \end{center}
\end{figure}

Ablations for the walls \textit{quadruped} are given in figure~\ref{fig:appendix:ablations:walls:ant}.

\end{document}